%% file: main.tex
\documentclass[10pt,twocolumn,letterpaper]{article}
\usepackage{colortbl}
\usepackage{caption}
\usepackage[utf8]{inputenc} 
\usepackage[T1]{fontenc}    
\usepackage{booktabs}       
\usepackage{amsfonts}       
\usepackage{nicefrac}       
\usepackage{microtype}      
\usepackage{times}
\usepackage{epsfig}
\usepackage{graphicx}
\usepackage{amsthm,amsmath,amssymb}
\usepackage{mathrsfs}
\usepackage{multirow}
\usepackage{wrapfig}
\usepackage{pifont}
\usepackage{rotating}
\usepackage{caption}
\usepackage{float}
\usepackage{stfloats}
\usepackage{appendix}  
\usepackage{cvpr}              

\input{preamble}

\usepackage{multirow}
\usepackage{pifont}
\usepackage{multirow}
\definecolor{cvprblue}{rgb}{0.21,0.49,0.74}
\usepackage[pagebackref,breaklinks,colorlinks,citecolor=cvprblue]{hyperref}
\newcommand{\zoubo}[1]{\textcolor{black}{#1}}
\newcommand\blfootnote[1]{%
  \begingroup
  \renewcommand\thefootnote{}\footnote{#1}%
  \addtocounter{footnote}{-1}%
  \endgroup
}
\def\paperID{10845} 
\def\confName{CVPR}
\def\confYear{2024}

\title{Teeth-SEG: An Efficient Instance Segmentation Framework for Orthodontic Treatment based on Multi-Scale Aggregation and Anthropic Prior Knowledge}
\author{Bo Zou *\\
Tsinghua University\\
Beijing, China\\
{\tt\small zoub21@mails.tsinghua.edu.cn}
\and
Shaofeng Wang *\\
Capital Medical Universty\\
Beijing, China\\
{\tt\small 2939108747@ccmu.edu.cn}
\and
Hao Liu\\
Tsinghua University\\
Beijing, China\\
{\tt\small liuh22@mails.tsinghua.edu.cn}
\and
Gaoyue Sun\\
Imperial College London\\
London, England\\
{\tt\small gaoyue.sun22@imperial.ac.uk}
\and
Yajie Wang\\
Tsinghua University, LargeV .Inc\\
Beijing, China\\
{\tt\small yj-wang18@mails.tsinghua.edu.cn}
\and
FeiFei Zuo\\
LargeV .Inc\\
Beijing, China\\
{\tt\small zuofeifei@largev.com}
\and
Chengbin Quan\\
Tsinghua University\\
Beijing, China\\
{\tt\small quancb@tsinghua.edu.cn}
\and
Youjian Zhao \dag\\
Tsinghua University, Zhongguancun Laboratory\\
Beijing, China\\
{\tt\small zhaoyoujian@tsinghua.edu.cn}
}

\begin{document}
\maketitle
\blfootnote{$^*$ Equal contribution, $^\dag$ Corresponding author \\
\indent\indent
Our code and dataset will be available at~\url{https://zoubo9034.github.io/TeethSEG/}}

\input{sec/0_abstract}   
\input{sec/1_intro}
\input{sec/2_relate}

\input{sec/3_method}

\input{sec/4_exp}
\input{sec/5_conclusion}
\section*{Acknowledge}
This work is supported in part by the Beijing Natural Science Foundation (No. L222024), the National Natural Science Foundation of China (No. 62394322), and the Beijing Hospitals Authority Clinical medicine Development of special funding support (No. ZLRK202330).

\clearpage
{
    \small
    \bibliographystyle{ieeenat_fullname}
    \bibliography{main}
}
\input{sec/6_append}


\end{document}

%% file: preamble.tex
\usepackage[dvipsnames]{xcolor}


%% file: sec/0_abstract.tex
\begin{abstract}
Teeth localization, segmentation, and labeling in 2D images have great potential in modern dentistry to enhance dental diagnostics, treatment planning, and population-based studies on oral health. However, general instance segmentation frameworks are incompetent due to 1) the subtle differences between some teeth' shapes (e.g., \zoubo{maxillary first premolar and second premolar}), 2) the teeth's position and shape variation across subjects, and 3) the presence of abnormalities in the dentition (e.g., caries and edentulism). To address these problems, we propose a ViT-based framework named TeethSEG, which consists of stacked Multi-Scale Aggregation (MSA) blocks and an Anthropic Prior Knowledge (APK) layer. Specifically, to compose the two modules, we design a unique permutation-based upscaler to ensure high efficiency while establishing clear segmentation boundaries with multi-head self/cross-gating layers to emphasize particular semantics meanwhile maintaining the divergence between token embeddings. Besides, we collect the first open-sourced intraoral image dataset IO150K, which comprises over 150k intraoral photos, and all photos are annotated by orthodontists using a human-machine hybrid algorithm. Experiments on IO150K demonstrate that our TeethSEG outperforms the state-of-the-art segmentation models on dental image segmentation. 
\end{abstract}

%% file: sec/1_intro.tex
\section{Introduction}
\label{sec:intro}

Malocclusion, caries, and periodontal disease are the three most common oral cavity diseases, especially the global incidence of malocclusion is a staggering $82.2\%$ \cite{cenzato2021prevalence}. Malocclusion is the misalignment between teeth, jaws, and craniofacial bones caused by genetic or environmental factors during a child's growth. According to the World Dental Federation (FDI), approximately 3.5 billion people worldwide suffer from malocclusion \cite{federation2020malocclusion}, which affects oral health, increasing the risk of caries, periodontal disease, and maxillofacial trauma, also affecting chewing, swallowing, breathing, and pronunciation. Orthodontic treatment is the primary means to cure malocclusion. It utilizes an orthodontic appliance to exert force on teeth in specific directions so that teeth can gradually move and finally achieve the goal of aligning the teeth, reaching the optimal occlusal function, and improving the appearance of the maxillofacial area.

The use of digital technology in oral orthodontics \cite{proffit2018contemporary,kim2021accuracy,payer2019integrating,nishimoto2019personal,zhong2019attention,park2019automated,moon2020much,hwang2020automated} has become a popular trend. One of the most widely used applications is integrating artificial intelligence technology to segment oral models and recognize tooth positions automatically. This integration significantly improves the efficiency of treatment plan design and reduces labor costs. Currently, all publicly available intraoral scan data (e.g., \cite{ben20233dteethseg, cui2022ctooth+}) and most teeth segmentation techniques \cite{xu20183d,cui2022ctooth,qiu2022darch,alsheghri2022semi,cui2021tsegnet} are in 3D space. Although 3D data provides more accurate maxillofacial structure recordings of patients, collecting 3D data is expensive as it requires costly equipment and trained professionals. Furthermore, processing 3D data is challenging and requires high computing resources, making it unsuitable for large-scale epidemiological screenings and self-inspections. In contrast, obtaining 2D data is relatively simple—a DSLR camera combined with a reflector can obtain standard intraoral dental images. With the improvement in the resolution of mobile phone cameras, individuals can also take their own clear intraoral photos. Dental practitioners can use 2D dental images to document various aspects of a patient's oral health, such as the alignment, count, color, and general condition of their teeth. By utilizing advanced 2D segmentation algorithms, these images can be analyzed to evaluate tooth crowding, occlusion status, anterior overbite/overjet, and midline alignment of the dental arch.

In recent years, transformer-based models \cite{zheng2021rethinking,strudel2021segmenter,chen2022vision,ranftl2021vision,anonymous2024llamaexcitor, bao2021beit,xie2021segformer,kirillov2023segment,anonymous2024videodistill,carion2020end, zou2023spaceclip}, have achieved remarkable success in computer vision, quickly dominating various tasks such as image classification, object detection, and semantic segmentation, surpassing traditional CNN models since they can better capture long-range dependencies and unify the modeling of different modalities. However, the teeth segmentation task is distinct from universal semantic segmentations, challenging the state-of-the-art transformer models. Firstly, unlike the apparent differences among object classes in common segmentation tasks, some teeth have similar appearances, such as maxillary first and second premolar. Accurate distinguishing between them requires a complete intraoral assessment. Secondly, due to the varying ages of patients, their teeth are at different stages of development and growth, resulting in different shapes and positions across subjects. Thirdly, caries and tooth loss, common in clinical orthodontic treatment, cause abnormalities in the dentition, which requires models to have strong generalization ability. Finally, to the best of our knowledge, there is no professional annotated 2D teeth segmentation dataset available to support training high-performance models.

To address the current situation, we create the first open-source 2D intraoral scan dataset IO150k, which consists of (1) Challenge80K, 80K rendered images generated from 1,800 3D scans sourced from 3D Teeth Scan Segmentation and Labeling Challenge 2023 \cite{ben20233dteethseg}, (2) Plaster70K, 70K images of 940 oral plaster models made before, during, and after taking the orthodontic treatment, and (3) RGB0.8K, 0.8K RGB standard intraoral photos taken before orthodontic treatment. This dataset has the following key properties: (1) Large: We have collected over 150K images (former dental datasets, e.g., \cite{wang2016benchmark,silva2018automatic,ajaz2013dental,li2021agmb}, have sizes around 0.1K to 3K ) that enable well-trained transformers that are usually more data-hungry than CNN models. (2) Diverse: We cover a wide range of dental malformations (e.g., crowded dentition and edentulism) to ensure the ability to generalize to clinical applications. (3) Professional: The data is annotated by multiple professional orthodontists using a human-machine hybrid algorithm, ensuring accurate tooth position recognition in complex instances. Please see Appendix A for dataset statistics.

Besides, we propose a novel transformer-based architecture designed against high-performance teeth segmentation named TeethSEG, which has two key components. The first one is Multi-Scale Aggregation Blocks (MSA) that effectively aggregate the visual semantics into trainable class embeddings of each tooth at different scales. The second one is the Anthropic Prior Knowledge Layer (APK), which imitates the principle of orthodontists to identify teeth, making the segmentation framework more interpretable and perform better, especially when tooth loss happens. Both modules are based on our specially designed multi-head self/cross-gating layers to emphasize valuable components in class embeddings while maintaining the divergence between them. In addition, most dense prediction frameworks \cite{chen2022vision,zheng2021rethinking,ranftl2021vision,liu2021swin,carion2020end} use transposed convolution to generate final segmentation maps. Some previous works, like \cite{strudel2021segmenter}, explore transformer-based decoders. However, they have trouble generating clear edges because the embedding sequences' length is much smaller than the final map size, resulting in mesh-like errors at the segmentation edges. In this paper, we explore replacing upsampling by compressing the intermediate feature dimensions to increase the sequence length, thereby enabling the encoder to learn to store rich local information in different parts of the patch embeddings. Our contributions are summarized as follows:
\begin{itemize}
\item We create IO150k, the largest open-source dataset that supports 2D dental segmentation. It covers a wide range of dental malformations and has professional annotations.
\item We propose TeethSEG with Multi-Scale Aggregation (MSA) blocks and the Anthropic Prior Knowledge (APK) layer, and the multi-head cross-gating mechanism and the permutation-based upscaler to form MSA and APK. 
\item Our experiments demonstrate that TeethSEG outperforms the state-of-the-art general-purpose segmentation models on dental image segmentation.
\end{itemize} 

%% file: sec/2_relate.tex
\section{Related Work}
\begin{figure*}[!t]
    \centering
    \includegraphics[width=0.9\textwidth]{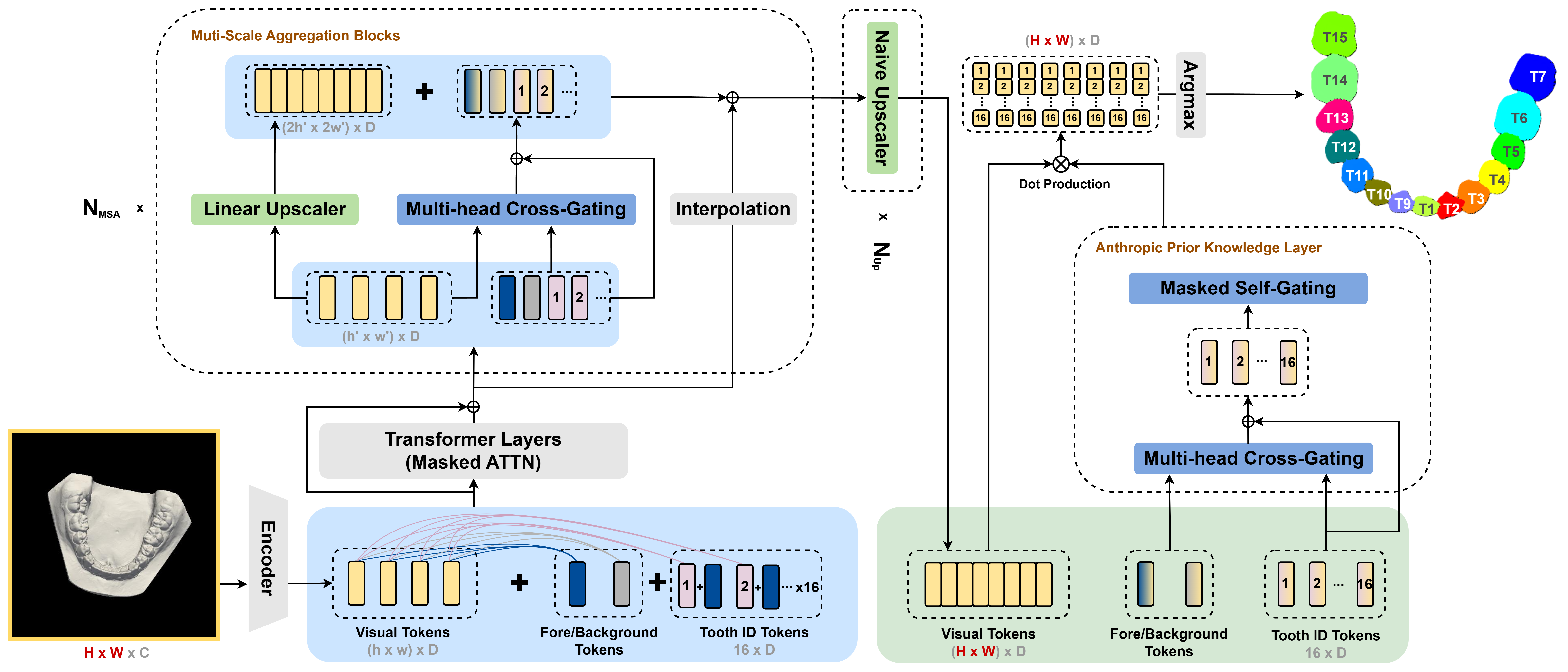}
    \caption{The overview of TeethSEG. We utilize a pretrained encoder to project an intraoral image into a sequence of visual tokens, and a set of trainable class tokens to predict segmentation masks. The multi-scale aggregation (MSA) blocks efficiently aggregate the visual information into class tokens, and the anthropic prior knowledge (APK) layer imposes human judgment into the mask prediction.}
    \label{fig:overview}
    \vspace{-12pt}
\end{figure*}
\textbf{Deep learning in Tooth Understanding.} 
Deep learning methods are increasingly used for 3D tooth segmentation. \cite{ tian2019automatic, zhang2020automatic, zanjani2019mask,leite2021artificial,im2022accuracy,zhao2021two,liu2022hierarchical}. Mask MCNet \cite{zanjani2019mask} combines the Monte Carlo Convolutional Network (MCCNet) with Mask R-CNN to locate each tooth object and segment all the tooth points inside the box. Graph convolutional network-based frameworks (GCN)  \cite{sun2020automatic, zhang2021tsgcnet, sun2020tooth} improve discriminative geometric feature learning for 3D dental model segmentation. TSegNet \cite{cui2021tsegnet} breaks down dental model segmentation into robust tooth centroid prediction and accurate individual tooth segmentation.  DArch \cite{qiu2022darch} proposes to estimate the dental arch and leverage the estimated dental arch to assist the proposal generation of tooth centroids. In summary, the previous method focuses on 3D teeth segmentation. We study instance segmentation for 2D intra-oral images, which lowers the data collection and annotation requirement and better supports large-scale epidemiological screenings and self-inspections.

\noindent
\textbf{Transformers in Dense Prediction.} In recent years, transformers have dominated various tasks. SETR \cite{zheng2021rethinking} is the first work to adopt ViT as the backbone and develop several CNN decoders for semantic segmentation. Segmenter \cite{strudel2021segmenter} also extends ViT to semantic segmentation but differs in that it equips a transformer-based decoder. DPT \cite{ranftl2021vision} further applies ViT to the monocular depth estimation task via a CNN decoder and yields remarkable improvements. Swin-transformer \cite{liu2021swin} proposes a shifted-window approach in computing self-attention. BeiT \cite{bao2021beit} applies masked image modeling as the pretraining tasks to strengthen the encoder. ViT-adapter \cite{chen2022vision} designs adapter blocks to inject inductive bias for ViTs to enhance performances in dense prediction. These works have achieved remarkable results on general segmentation datasets. However, the teeth segmentation task is distinct from universal semantic segmentations, challenging the state-of-the-art transformer models.

%% file: sec/3_method.tex
\section{Methodology}
\label{method}
The primary goal of TeethSEG is to better identify the categories of each individual tooth rather than just distinguishing between tooth areas and background \zoubo{(gingiva)} areas.
Meanwhile, we make efforts to capture clear segmentation edges with a pure transformer architecture. We choose the multi-model pretrained CLIP encoder as our backbone as its effectiveness has been demonstrated in many downstream tasks. Additionally, its ability to align images with text makes it a strong foundation for expanding TeethSEG into a multi-modal diagnostic model in the future. In Sec \ref{overall}, we first introduce how to generate segmentation masks based on the pretrained encoder. Then, in Sec \ref{gate} and Sec \ref{upscaler}, we introduce the multi-head cross/self-gating mechanism and the permutation-based upscalers (including a naive upscaler and a linear upscaler) that make up our Multi-Scale Aggregation Blocks (MSA) and the Anthropic Prior Knowledge Layer (APK). Finally, in Sec \ref{MSA} and Sec \ref{APK}, we present the details of MSA and APK that are specifically designed for teeth segmentation.

\subsection{Overall architecture}
\label{overall}
An image $X \in \mathbb{R}^{H \times W \times C}$ is encoded into a sequence of visual tokens $x = \left[x_{1}, \dots, x_{N} \right] \in \mathbb{R}^{N \times D}$ by a pretrained encoder, where $N = hw = HW / P^{2}$ is the number of visual tokens, $\left(P, P \right)$ is the patch size, and $D$ is the dimension of embeddings. Visual tokens $x$ carry rich visual information in the image. We introduce a set of 18 trainable class embeddings to gather the features of the foreground (teeth region), background (gingiva), and each individual tooth. They are divided into foreground/background tokens $CLS_{fb} \in \mathbb{R}^{2 \times D}$ and tooth ID tokens $CLS_{th} \in \mathbb{R}^{16 \times D}$. 
Before the following computations, as shown in Fig \ref{fig:overview}, we first add the embedding of the foreground to each tooth ID token in $CLS_{th}$ since all tooth areas should be included in the foreground. After that, We use a $M$-layer transformer with masked attention to fuse visual tokens and learnable tokens. As shown at the bottom of Fig \ref{fig:overview}, we apply the attention mask within learnable tokens and only allow visual tokens $x$ to update each token in $CLS_{fb}$ and $CLS_{th}$ at this stage because we want to maximize the dissimilarity within tooth ID tokens $CLS_{th}$. In this way, we can mitigate the difficulties in distinguishing similar tooth categories. To better merge multi-scale visual semantics into learnable tokens, we utilize MSA blocks in Sec \ref{MSA}, which takes shallow fused $x$, $CLS_{fb}$, and $CLS_{th}$ as input to perform deeper feature interaction under different receptive fields. Then, we up-sample the intermediate visual tokens $x'$ to $x' \in \mathbb{R}^{ ( H \times W ) \times D}$ that match the size of the input image $X$ by the permutation-based upscaler in Sec \ref{upscaler}. Finally, we enable the interactions within learnable tokens under the instruction of human prior knowledge by the APK layer in Sec \ref{APK}. The class masks of each tooth are generated by computing the softmax of the scalar product between $x'$ and tooth ID tokens $CLS_{th}$ as follows:
\begin{small}
\begin{equation}
    score^{th} = {\rm{softmax}}\left(\frac{ x' CLS_{th}^{\rm{T}} }{ \sqrt{D} }\right),
    \label{equation1}
\end{equation}
\end{small}
where $score^{th} \in \mathbb{R}^{(H \times W) \times 16}$ is the pixel-wise class score. The $\sqrt{D}$ in the denominator prevents numerical overflow and stabilizes the training. Similarly, the class masks of the foreground and the background are formulated as follows:
\begin{small}
\begin{equation}
    score^{fb} = {\rm{softmax}}\left(\frac{ x' CLS_{fb}^{\rm{T}} }{ \sqrt{D} }\right).
    \label{equation2}
\end{equation}
\end{small}

Our model is trained end-to-end with a per-pixel cross-entropy loss consisting of two parts:
\begin{small}
\begin{equation}
    \mathcal{L}_{th} = -\frac{1}{H W} \sum_{i=1}^{H W} y_i \log(score^{th}_{i}),
    \label{equation3}
\end{equation}
\begin{equation}
    \mathcal{L}_{fb} = -\frac{1}{H W} \sum_{i=1}^{H W} y_i \log(score^{fb}_{i}),
    \label{equation4}
\end{equation}
\end{small}
where $y_{i}$ is the label of the $i$-th pixel.

\subsection{Multi-Head Cross/Self-Gating Mechanism}
\label{gate}
\begin{figure}[t]
    \centering
    \includegraphics[width=0.4\textwidth]{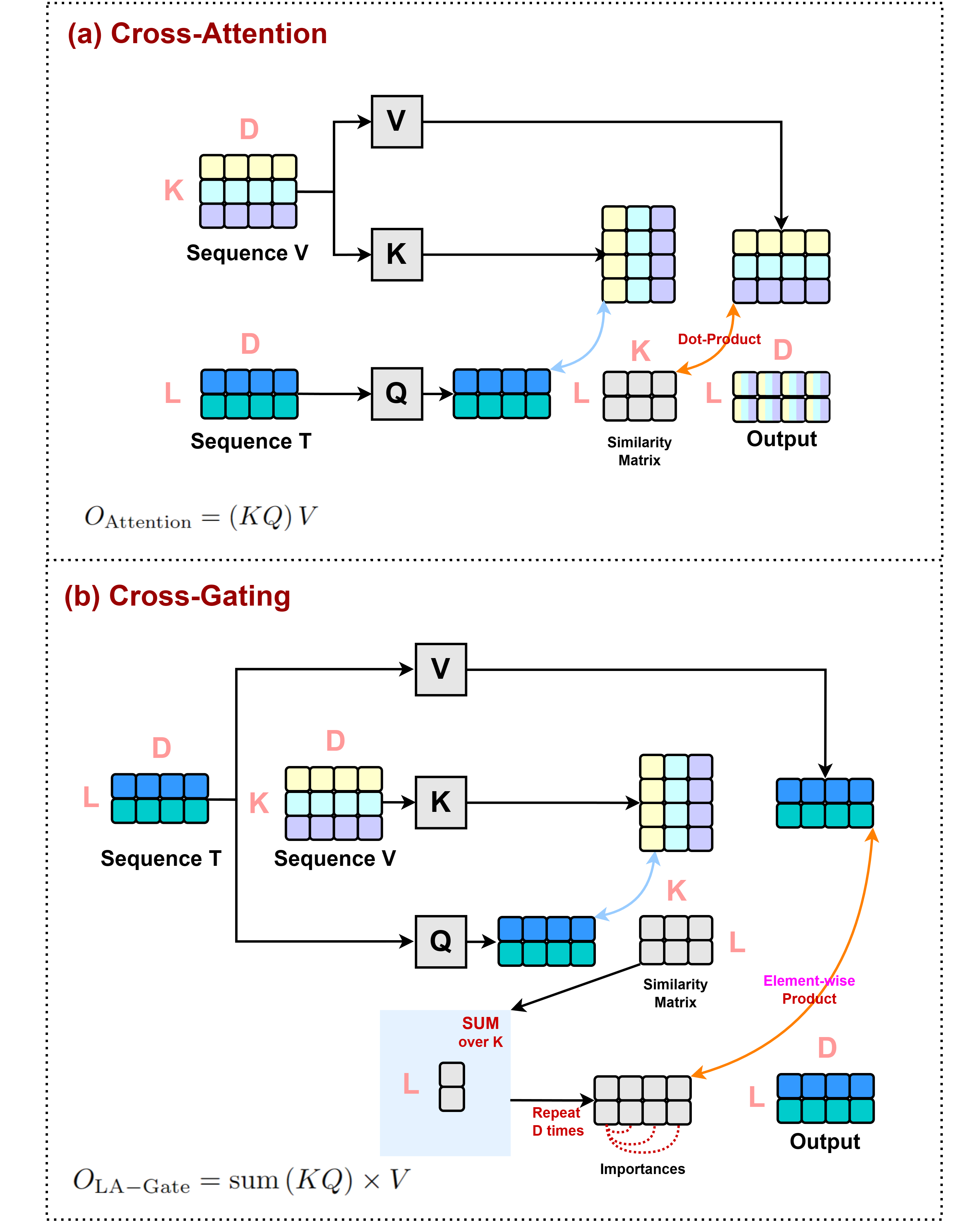}
    \caption{Illustrations of Cross-Attention and our Cross-Gating mechanisms}
    \label{fig:gate}
    \vspace{-12pt}
\end{figure}

We introduce a reusable unit termed cross(self)-gating mechanism for MSA and APK, which takes two arbitrary sub-sequences $V \in \mathbb{R}^{K \times D}$ and $T \in \mathbb{R}^{L \times D}$ as input and performs more efficient feature interactions than commonly used cross-attention after the earlier fusions in the transformer, by exciting or depressing the components of $T$ according to their similarities with $V$. For a better understanding of the cross-gating Mechanism, we illustrate it and the cross-attention in Fig \ref{fig:gate}.

There are two key operations of our cross-gating. (1) When $K$, the length of $V$, is larger than 1, we sum the similarity matrix $S \in \mathbb{R}^{L\times K}$ over $K$ to form a vector $I \in \mathbb{R}^{L}$ of importance for token embeddings in $T$. $K$ functions like the number of multi-heads in the attention mechanism, and every token embedding in $V$ will partially dictate the importance of each token embedding in $T$. (2) We expand the importance vector $I$ (repeat D times) to match the shape of $T$. Then, we apply the element-wise product rather than the dot product on the importance matrix and the linear-projected $T$ ($Keys$).The whole process is formulated as follows:
\begin{small}
\begin{equation}
\begin{aligned}
Keys = W_{\rm{k}} \left(V \right), 
Querys = W_{\rm{q}} \left(T \right),      
Values = W_{\rm{v}} \left(T \right)
\label{equation5}
\end{aligned}
\end{equation}
\begin{equation}
I=\mathrm{repeat} \left( \mathrm{sum} \left(\frac{Querys\cdot Keys}{\left\|Querys\right\|\times\left\|Keys\right\|} \right)\right), 
\label{equation6}
\end{equation}

\begin{equation}
Output = I \odot Values, \label{equation7}
\end{equation}
\end{small}
\noindent
where $\odot$ denotes element-wise production. In practice, we perform a standard multi-head attention setting \cite{vaswani2017attention} on $W_{\rm{k}}$, $W_{\rm{q}}$, $W_{\rm{v}}$, and concatenate outputs of each head.

The most significant characteristic of cross-gating is it can better maintain local diversity within $T$. In Fig \ref{fig:gate} (a), cross-attention's output displays rows in mixed colors, representing the weighted sum of token embeddings. Consequently, it demonstrates a more global attribute. By contrast, we maintain the uniqueness of colors for cross-gating in Fig \ref{fig:gate} (b) because each token embedding is only multiplied by their importance, which is a scalar. This feature is crucial for TeethSEG because the divergence in tooth ID tokens allows us to better distinguish between tooth categories with high similarity. Besides, the interactions brought by commonly used cross-attention (updating embeddings by the weighted sum) can be covered in the previous $M$-layers transformer when it is applied in MSA and APK since $T$ and $V$ are coming from the same output sequence.

\subsection{Permutation-based Upscaler}
\label{upscaler}
We upscale the intermediate visual token sequence $x' \in \mathbb{R}^{(h\times w) \times D}$ to $x' \in \mathbb{R}^{ (2h \times 2w) \times (D/4)}$ by equally dividing every token embedding with the dimension of $\mathbb{R}^{D}$ in $x'$ into a sequence with the shape of $\mathbb{R}^{4\times (D/4)}$ and permute them.
We name this process as \textbf{naive upscaler} (on the top of Fig \ref{fig:overview}) since it simply increases the spatial dimensions by compressing the dimension of feature embedding.

However, this naive upscaler performs significantly better than the bilinear interpolation. Applying bilinear interpolation on $x'$ as done in previous techniques \cite{strudel2021segmenter} does not generate clear segmentation edges. This is because the local information of the interpolated feature map is highly similar, which causes mesh-like errors at the segmentation edge. Besides, the naive upscaler can impose the image encoder to maintain rich local information in its visual tokens.

\begin{figure*}[!t]
    \centering
    \includegraphics[width=0.8\textwidth]{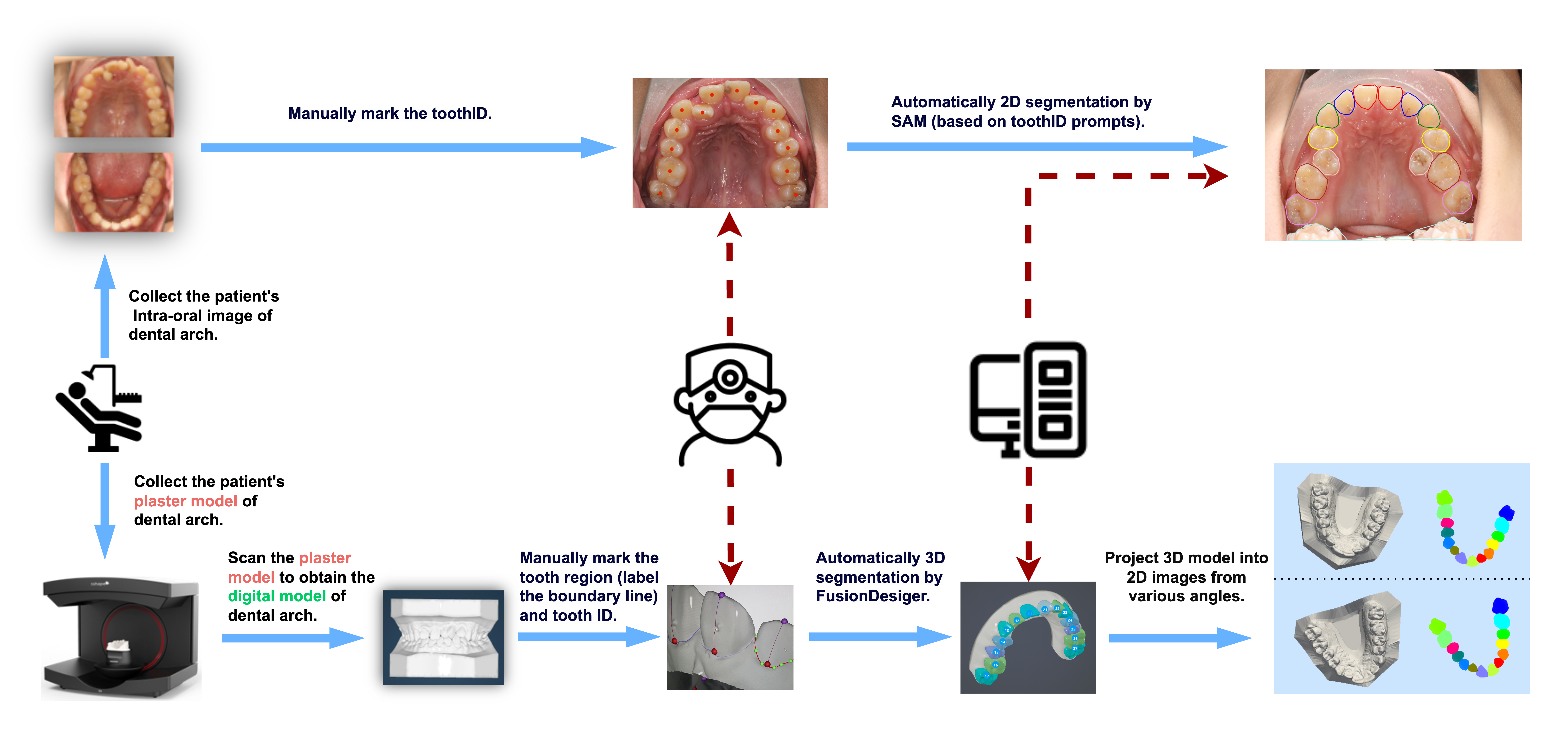}
    \caption{Illustration of our human-machine hybrid data annotation process.}
    \label{fig:annotation}
\end{figure*}

\subsection{Multi-Scale Aggregation Block}
\label{MSA}
The $N_{\rm{MSA}}$ stacked multi-scale aggregation blocks (MSA) are designed to take in shallow fused $x$, $CLS_{fb}$, and $CLS_{th}$ and perform deeper feature interaction under different receptive fields. As shown on the top of Fig \ref{fig:overview}, each MSA block first uses a cross-gating layer to enhance important semantics in $CLS_{th}$ and $CLS_{fb}$ according to the intermediate visual tokens $x' \in \mathbb{R}^{(h' \times w' \times D)}$, where $h' = 2^k \times h$, $w' = 2^k \times w$, and $k \in [0, N_{\rm{MSA}}]$. (In Fig \ref{fig:gate}, $V$ stands for visual tokens $x'$, $T$ stands for $CLS_{fb}$ and $CLS_{th}$). Then, it upsamples $x'$ by what we call \textbf{linear upscaler} (on the left of Fig \ref{fig:overview}), a combination of a linear projection layer $W_{\rm{U}}$ and our naive upscaler, where $W_{\rm{U}} \in \mathbb{R}^{(D/4 \times D)}$ is used to maintain the embedding dimension $D$. By stacking $N_{\rm{MAS}}$ MSA blocks, we can refine the class token embeddings according to multi-scale visual semantics. Additionally, we apply skip connections between each MSA block. The skip connection utilizes the bilinear interpolation to upsample $x$.

\subsection{Anthropic Prior Knowledge Layer}
\label{APK}
We propose a scalable modular APK to introduce human prior knowledge into the segmentation process. In this paper, we summarize three prior rules based on the annotation experience of orthodontists in complex situations such as tooth loss or abnormal tooth counts in the dental arch. The first rule states that the region requiring tooth ID labeling should not be in the background area, such as the gingiva. The second rule emphasizes the importance of considering the morphological structure of the adjacent teeth on the left and right of each tooth. Finally, the third rule suggests investigating the morphological structure of the contralateral teeth of each tooth based on the symmetry of tooth growth.

To comply with these rules, APK first utilizes a cross-gating layer, which takes $CLS_{fb}$ and $CLS_{th}$ as input (In Fig \ref{fig:gate}, $V$ represents $CLS_{fb}$ and $T$ represents $CLS_{th}$), to emphasize the knowledge of foreground in tooth ID tokens. Then, it sends only the processed tooth ID tokens $CLS_{th}$ into a masked self-gating layer, which means $V$ and $T$ in Equation \ref{equation5} are the same sequence. This layer has an attention mask on the important matrix $I$ in Equation \ref{equation6}, which only enables interaction between adjacent and contralateral teeth, to meet the second and third rules' requirements. The attention mask is visualized in \zoubo{Appendix B}.

%% file: sec/4_exp.tex
\section{Experiments}
\label{exp}
\input{table/iid}

\subsection{Dataset annotation and processing.}
The data annotation, i.e., teeth segmentation and labeling, was performed in collaboration with 4 orthodontists with over 6 years of clinical training experience. The orthodontists were trained in the FDI tooth notation method \cite{keiser1971federation}, as well as how to use the annotation software and adhere to the annotation standard. The annotation standard requires each orthodontist to independently annotate all visible deciduous and permanent teeth in each type of intraoral (3D scans of plaster models and 2D RGB photos) data within 7 days. After three weeks, they review all annotations to correct any errors and missed tooth labels. Note, for 2D intra-oral photos, the annotations include teeth with exposed coronal parts or visible residual crowns and roots in the photos but exclude teeth reflected in the reflector for intraoral photography. The detailed process is depicted in Fig \ref{fig:annotation}. 

Our approach to reducing labor costs involves a combination of human and machine annotations. To achieve this, we use FusionAnalyser \cite{zhang2023Fusion}, a dental model analysis tool, for 3D scans (on the bottom of Fig \ref{fig:annotation}). In this process, orthodontists draw the boundary line and identify the corresponding tooth ID for each tooth region. The software then automatically generates 3D segmentation for each tooth. For 2D photos (on the top of Fig \ref{fig:annotation}), we first ask orthodontists to label the central point of each tooth class. We then use SAM \cite{kirillov2023segment}, an open-source image segmentation framework, to generate segmentation masks based on the human-labeled tooth centers. Finally, we ask orthodontists to verify all auto-generated segmentations for both data types. This process ensures the accuracy of the final segmentation. Finally, we rotate the 3D models and project them onto 2D images with labels in various angles. Our method significantly increases the dataset's sample richness while minimizing sample collection costs and annotation costs. Besides, training on a multi-angle plaster cast also helps to improve the model's tolerance for low-quality intra-oral shots (camera angle skew).

\subsection{Experimental Setup}
\textbf{Competing Methods.} We compare our approach with the state-of-the-art methods (i.e., DeepLabV3 \cite{chen2017deeplab}, Segmenter \cite{strudel2021segmenter}, Segformer \cite{xie2021segformer}, Swin-transformer \cite{liu2021swin}, BeiT \cite{bao2021beit}, and ViT-adapter \cite{chen2022vision}) of 2D instance segmentation. DeepLabV3 is a powerful DilatedFCN-based model with atrous spatial pyramid pooling introducing rich multi-scale information. Segmenter, which uses a masked transformer to generate segmentation masks, is an earlier attempt that brings the vision transformer (ViT) into the field of semantic segmentation. Segmentor comes up with using overlapped image patches 
to increase local continuity for ViT-based models and uses deep-wise convolutions to replace the positional embedding. Swin-transformer proposes a shifted-window approach in computing self-attention, increasing token embeddings' scale while reducing overhead. BeiT applies masked image modeling (MIM), a token-level autoregression, as the pretraining tasks to strengthen the encoder. ViT-adapter designs adapter block to inject inductive bias for ViTs to enhance performances in dense prediction.

\noindent
\textbf{Implementation Details.}
Our IO150K contains three parts: (1) Challenge80K, 80K rendered images from 3D scans provided by 3D Teeth Scan Segmentation and Labeling Challenge 2023 \cite{ben20233dteethseg}, (2) Plaster70K, 70K images of oral plaster models, and (3) RGB0.8K, 0.8K RGB standard intra-oral photos. Each part has been individually divided into training, validation, and testing splits (please see \zoubo{Appendix A} for details). Although all three parts of IO150K support separate training and testing for future studies, in this paper, we first pretrain models on the training split of Challenge80K and test on Challenge80K and Plaster70K testing splits (denoted as \textbf{i.i.d. test} (independent and identically distributed), \textbf{o.o.d. test} (out of the distribution)). then finetune and test models on RGB0.8K (denoted as \textbf{RGB test}). This is because the data in Challenge80K is general tooth data that matches the real-world distribution, while the Plaster80K and RGB0.8K we collected are from patients who accept the orthodontic examination. Compared with Challenge80K, their samples have more abnormalities and higher complexity. We hope the model trained on general data can adapt to the needs of orthodontic diagnosis (o.o.d. test), and the models trained on a large number of 3D model projections can be transferred to the use of analyzing RGB intra-oral photos for early screening (RGB test).

We report the results of TeethSEG with pretrained CLIP-L/14@336 as the encoder. The embedding dim $D=768$, the number of layers in the masked transformer $M=3$, the number of stacked MSA blocks $N_{\rm{MSA}} = 3$, and the number of stacked naive upscalers $N_{\rm{up}} = 2$.  


\subsection{Comparison with Competing Methods}
\textbf{I.I.D. Test Results.} The overall detection and segmentation results are presented in Table \ref{table:iid}, and we compare these competing methods in the Intersection over Union (IoU) of each tooth class (denoted as T1 to T16. The pre-defined order of classes is shown on the top of Fig \ref{fig:overview}. NaN means the corresponding class is not shown in our test split.) and the average over all classes (mIoU). The table shows that our TeehSEG achieves the best segmentation performance in each tooth class and improves overall segmentation performance by 4\% compared with the state-of-the-art methods. From the table, we can see that for the i.i.d. test, competing methods have similar performance on teeth with a large area (i.e., T6, T7, T14, T15) and have significant performance differences on smaller teeth (i.e., T1, T2, T9, T10). In particular, methods targeting capturing multi-scale objects (i.e., DeepLab-v3, Segformer, ViT-Adapter) have better performances on smaller teeth. Our TeethSEG uses MSA Blocks to capture multi-scale information and improves 4\% to 5\% IoU performance on T1, T2, T9, and T10.

\input{table/ood_and_rgb}

\input{table/show}
\input{table/ablation}
\noindent
\textbf{O.O.D. Test Results.} We compare the o.o.d. performance of TeethSEG with competing methods in Table \ref{tab:ood&rgb} and find that our method outperforms the state-of-the-art methods with 3\% mIoU, which shows TeethSEG's generalize ability on data ad-hoc to orthodontic treatment. We also visualize randomly picked 4 segmentation predictions of some methods and highlight the incorrect parts in Table \ref{table:case}. From this table, we can find competing methods have different degrees of errors in dealing with complex situations such as missing teeth or irregular tooth arrangements. Interestingly, the methods that perform well in the i.i.d. test still obtain clear tooth segmentation boundaries in this test (i.e., correctly distinguishing between tooth areas and background areas). However, they all have the problem of incorrectly categorizing some teeth as belonging to other tooth categories (in the ground truth, we assigned a unique color to each tooth ID). By contrast, our framework can better identify tooth IDs by incorporating relevant dental arch information, with the help of human prior knowledge via the APK layer in TeethSEG. Please see Appendix C for the full visualization.

\begin{figure}[h]
    \centering
    \includegraphics[width=0.45\textwidth]{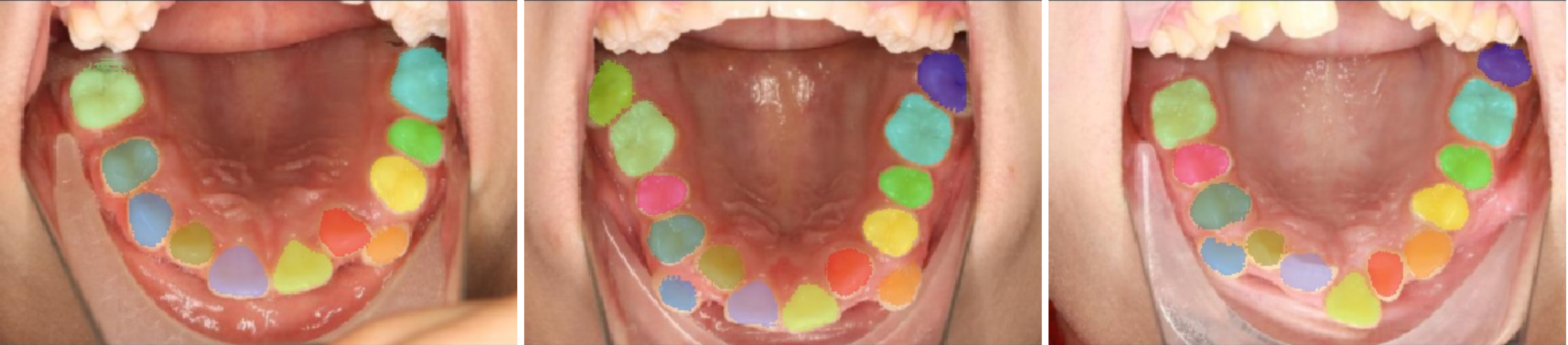}
    \caption{Examples of TeethSEG's segmentation results on IO150K RGB test split.}
    \label{fig:rgb}
\end{figure}

\noindent
\textbf{RGB Test Results.} We finetune the models on our RGB0.8K to show the pretrained knowledge on plaster models can be transferred into the RGB domain. Table \ref{tab:ood&rgb} reports the performances of TeethSEG with competing methods. It shows that TeethSEG brings 6\% performance boost, which demonstrates the generalization ability of our framework. Fig \ref{fig:rgb} visualizes several randomly picked prediction results of TeethSEG on intra-oral photos from patients before receiving orthodontic treatment. We find that even in cases of obvious dental arch abnormalities, TeethSEG can still accurately segment the tooth area and identify the correct tooth ID. Please see Appendix C for the visual comparison with other methods.

\begin{figure}[t]
    \centering
    \includegraphics[width=0.5\textwidth]{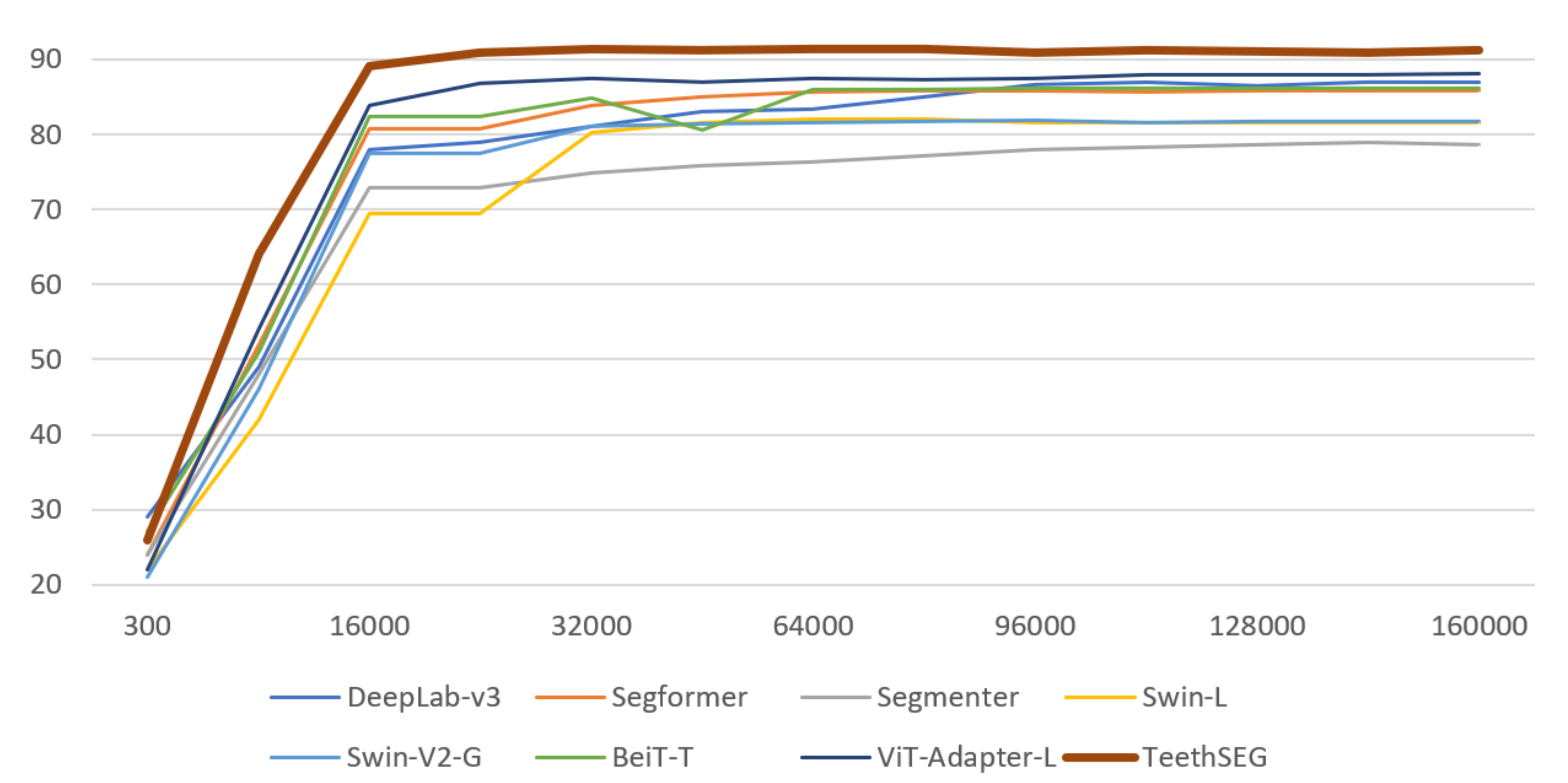}
    \caption{The trend of mIoU changes during the training process.}
    \label{fig:speed}
    \vspace{-12pt}
\end{figure}

\noindent
\textbf{Comparison on Training Speed.}
Due to the specialized design introduced by TeethSEG for tooth segmentation, its training speed is higher than competing methods. In Fig \ref{fig:speed}, we visualize the change of mIoU on Challenge80K during the pretraining for all methods.

\subsection{Ablation}


\begin{figure}[h]
    \centering
    \includegraphics[width=0.45\textwidth]{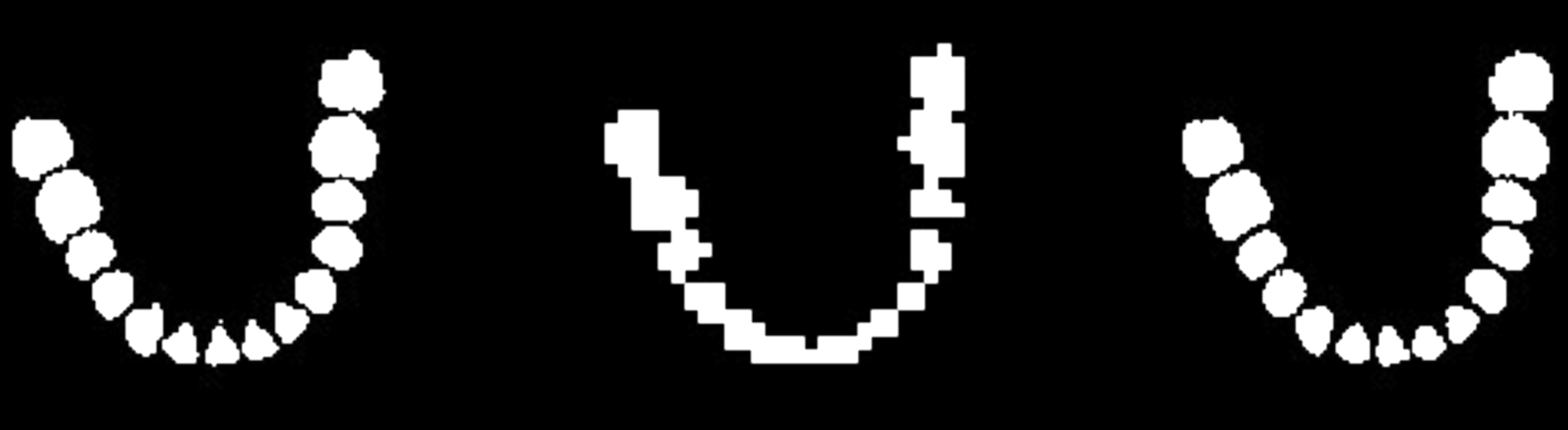}
    \caption{Comparison of Bilinear Interpolation to Permutation-based Upscaler. (left) Ground Truth, (middle) Bilinear Interpolation, (right) Permutation-based Upscaler.}
    \label{fig:background}
\end{figure}

\noindent
\textbf{Permutation-based Upscaler vs. Bilinear Interpolation.}
Previous transformer-only decoders used bilinear interpolation for scaling the intermediate feature map to match the size of the output, causing errors at segmentation edges. Besides, the interpolated enlarged image does not introduce new information to local areas, which prohibits the model from learning multi-scale semantics during training. In Fig \ref{fig:background}, we visualize the background segmentation result generated by using $CLS_{fb}$ in Sec. \ref{overall} and compare it with the result from a variant of replacing all linear upscalers and naive upscalers with bilinear interpolation. We also quantitate the performance difference in Table \ref{table:effectiveness}.

\noindent
\textbf{Effectiveness of each component.} 
To verify the effectiveness of our permutation-based upscalers, cross-gating mechanism, MSA blocks, and the APK layer, we design six variants and report their performances in Table \ref{table:effectiveness}. (a) We replace all linear upscalers and naive upscalers with bilinear interpolation. (b) We replace our proposed cross/self-gating with the standard cross/self-attention in the MSA blocks and the APK layer. In (c), (d), and (e), we explore the influence of removing the MSA bocks or the APK layer, as well as removing them all. (f) is our best variant with all specially designed modules.

\noindent
\textbf{More Ablations.} We validate the influence of the scale of the image encoder, the resolution of the input images, and the reasonable choice of hyper-parameters in \zoubo{Appendix D}.


%% file: table/iid.tex
\begin{table*}[!t]
    \centering
    \resizebox{0.8\textwidth}{!}{%
    \begin{tabular}{l c ccccccccc cccccccc c}
    \toprule
    Method          &Epoch  &T1     &T2     &T3     &T4     &T5     &T6     &T7     &T8     &T9     &T10    &T11    &T12    &T13    &T14    &T15    &T16    &mIoU\\
    \midrule
    DeepLab-v3      &30     &0.85   &0.85   &0.85   &0.89   &0.91   &0.96   &0.87   &NaN    &0.85   &0.83   &0.89   &0.91   &0.83   &0.85   &0.82   &NaN    &0.87\\
    Segformer       &30     &0.86   &0.85   &0.84   &0.86   &0.91   &0.93   &0.87   &NaN    &0.87   &0.84   &0.88   &0.91   &0.77   &0.77   &0.72   &NaN    &0.85\\
    Segmenter       &30     &0.78   &0.77   &0.74   &0.68   &0.64   &0.77   &0.75   &NaN    &0.78   &0.76   &0.81   &0.86   &0.71   &0.75   &0.69   &NaN    &0.75\\
    Swin-L          &30     &0.81   &0.79   &0.77   &0.78   &0.82   &0.89   &0.85   &NaN    &0.82   &0.80   &0.83   &0.87   &0.74   &0.75   &0.70   &NaN    &0.80\\
    SwinV2-G        &30     &0.81   &0.79   &0.77   &0.84   &0.84   &0.90   &0.85   &NaN    &0.81   &0.79   &0.83   &0.87   &0.74   &0.76   &0.71   &NaN    &0.80\\
    BeiT-B          &30     &0.79   &0.76   &0.76   &0.78   &0.84   &0.90   &0.84   &NaN    &0.80   &0.77   &0.81   &0.85   &0.77   &0.80   &0.74   &NaN    &0.80\\
    ViT-Adapter-L   &30     &0.87   &0.87   &0.84   &0.86   &0.90   &0.93   &0.87   &NaN    &0.88   &0.87   &0.89   &0.87   &0.82   &0.87   &0.82   &NaN    &0.87\\
    \midrule
    TeethSEG        &5      &0.92   &0.92   &0.84	&0.89   &0.94   &0.96   &0.89   &NaN    &0.92	&0.92	&0.93	&0.95	&x0.85	&0.87	&0.82   &NaN    &0.91\\
    \bottomrule
    \end{tabular}
    }   
    \caption{Results (mIoU) compared with SOTA methods on the IO150K independent and identically distributed (i.i.d.) test splits}
    \label{table:iid}
    \vspace{-12pt}
\end{table*}

%% file: table/ood_and_rgb.tex
\begin{table}[!t]
    \centering
    \tiny
    \resizebox{0.65\linewidth}{!}{%
    \begin{tabular}{l|c|cc}
    \toprule
    \multirow{2}{*}{Method}     &\multirow{2}{*}{Epoch}    &\multicolumn{2}{c}{mIoU} \\
       &                       &o.o.d.     &RGB \\
    \midrule
    DeepLab-v3      &30         &0.80       &0.80\\
    Segformer       &30         &0.81       &0.69\\
    Segmenter       &30         &0.68       &0.55\\
    Swin-L          &30         &0.64       &0.48\\    
    SwinV2-G        &30         &0.60       &0.46\\
    BeiT-B          &30         &0.78       &0.47\\
    ViT-Adapter-L   &30         &0.79       &0.85\\
    \midrule
    TeethSEG        &5          &0.84       &0.91\\
    \bottomrule
    \end{tabular}
    }
    \caption{Tooth segmentation results (mIoU) on the IO150K out-of-the-distribution (o.o.d.) test splits and RGB test split. Please see the Appendix for the IoU of each tooth ID.}
    \label{tab:ood&rgb}
    \vspace{-12pt}
\end{table}

%% file: table/show.tex
\begin{table*}
    \begin{minipage}{1\textwidth} 
    
    \begin{minipage}{0.15\textwidth} 
    \centering
    DeepLab-v3
    \end{minipage}
    \begin{minipage}{0.15\textwidth} 
    \centering
    BeiT-B
    \end{minipage}
    \begin{minipage}{0.15\textwidth} 
    \centering
    SwinV2-G
    \end{minipage}
    \begin{minipage}{0.15\textwidth} 
    \centering
    ViT-Adapter-L
    \end{minipage}
    \begin{minipage}{0.15\textwidth} 
    \centering
    TeethSEG
    (Ours)
    \end{minipage}
    \begin{minipage}{0.15\textwidth} 
    \centering
    Ground Truth
    \end{minipage}
    
    \end{minipage}
    \par
    \begin{minipage}{1\textwidth} 
    
    \begin{minipage}{0.15\textwidth} 
    \includegraphics[width=0.9\textwidth,height=0.7\textwidth]{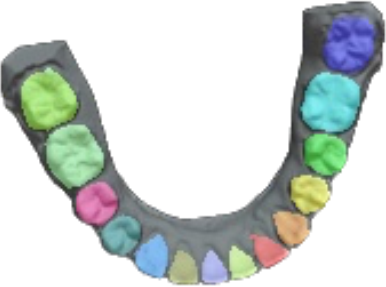}  
    \end{minipage}
    \begin{minipage}{0.15\textwidth} 
    \includegraphics[width=0.9\textwidth,height=0.7\textwidth]{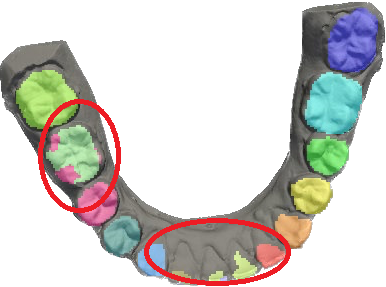}  
    \end{minipage}
    \begin{minipage}{0.15\textwidth} 
    \includegraphics[width=0.9\textwidth,height=0.7\textwidth]{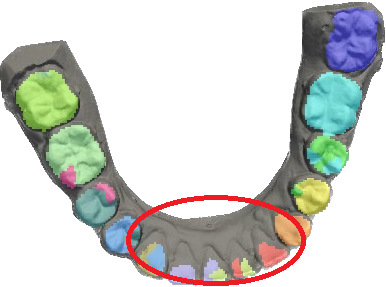}  
    \end{minipage}
    \begin{minipage}{0.15\textwidth} 
    \includegraphics[width=0.9\textwidth,height=0.7\textwidth]{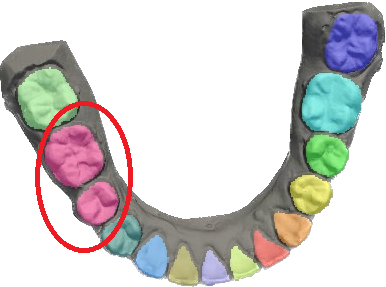}  
    \end{minipage}
    \begin{minipage}{0.15\textwidth} 
    \includegraphics[width=0.9\textwidth,height=0.7\textwidth]{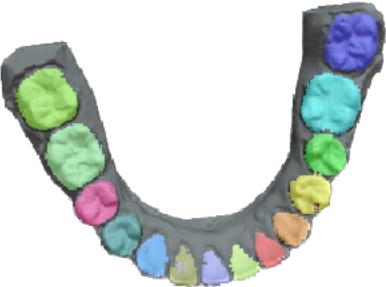}  
    \end{minipage}
    \begin{minipage}{0.15\textwidth} 
    \includegraphics[width=0.9\textwidth,height=0.7\textwidth]{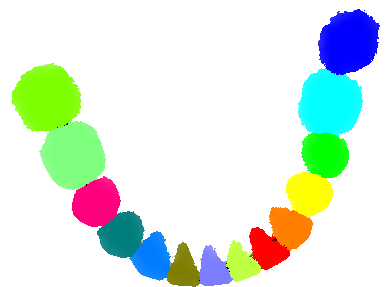}  
    \end{minipage}
    
    \end{minipage}
    \par
    \begin{minipage}{1\textwidth} 
    
    \begin{minipage}{0.15\textwidth} 
    \includegraphics[width=0.9\textwidth,height=0.7\textwidth]{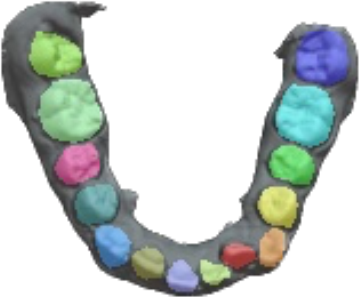}  
    \end{minipage}
    \begin{minipage}{0.15\textwidth} 
    \includegraphics[width=0.9\textwidth,height=0.7\textwidth]{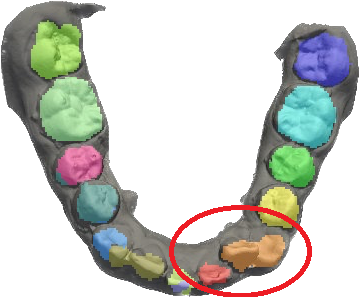}  
    \end{minipage}
    \begin{minipage}{0.15\textwidth} 
    \includegraphics[width=0.9\textwidth,height=0.7\textwidth]{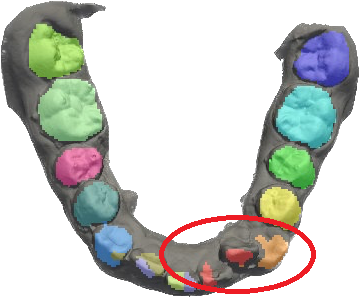}  
    \end{minipage}
    \begin{minipage}{0.15\textwidth} 
    \includegraphics[width=0.9\textwidth,height=0.7\textwidth]{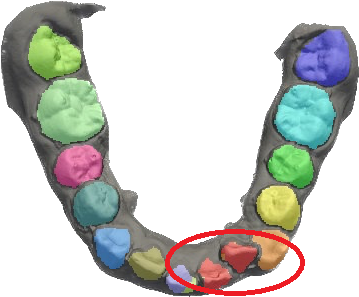}  
    \end{minipage}
    \begin{minipage}{0.15\textwidth} 
    \includegraphics[width=0.9\textwidth,height=0.7\textwidth]{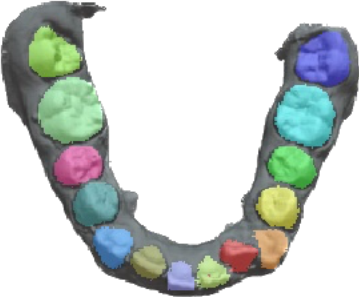}  
    \end{minipage}
    \begin{minipage}{0.15\textwidth} 
    \includegraphics[width=0.9\textwidth,height=0.7\textwidth]{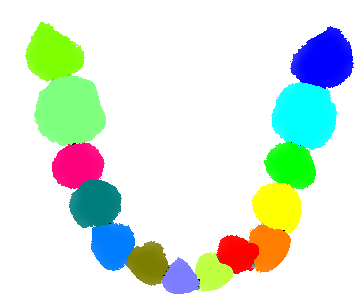}  
    \end{minipage}

    \end{minipage}
    \par
    \begin{minipage}{1\textwidth} 
    
    \begin{minipage}{0.15\textwidth} 
    \includegraphics[width=0.9\textwidth,height=0.7\textwidth]{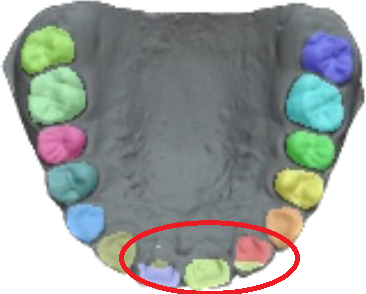}  
    \end{minipage}
    \begin{minipage}{0.15\textwidth} 
    \includegraphics[width=0.9\textwidth,height=0.7\textwidth]{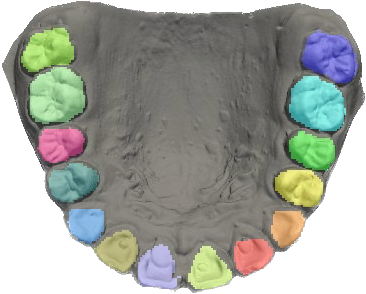}  
    \end{minipage}
    \begin{minipage}{0.15\textwidth} 
    \includegraphics[width=0.9\textwidth,height=0.7\textwidth]{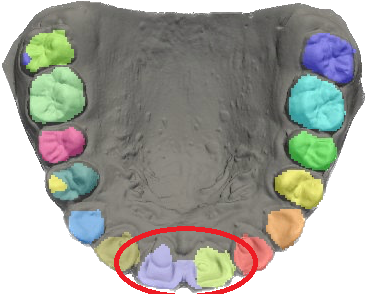}  
    \end{minipage}
    \begin{minipage}{0.15\textwidth} 
    \includegraphics[width=0.9\textwidth,height=0.7\textwidth]{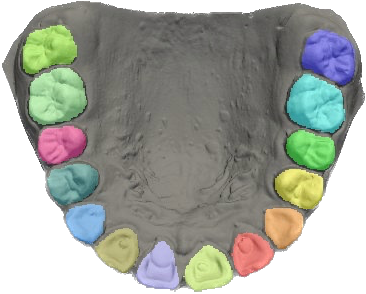}  
    \end{minipage}
    \begin{minipage}{0.15\textwidth} 
    \includegraphics[width=0.9\textwidth,height=0.7\textwidth]{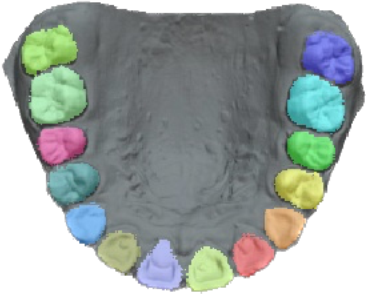}  
    \end{minipage}
    \begin{minipage}{0.15\textwidth} 
    \includegraphics[width=0.9\textwidth,height=0.7\textwidth]{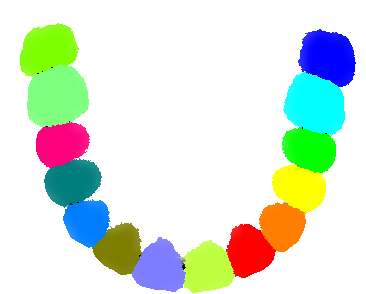}  
    \end{minipage}   
    
    \end{minipage}
    \par
    \begin{minipage}{1\textwidth} 
    
    \begin{minipage}{0.15\textwidth} 
    \includegraphics[width=0.9\textwidth,height=0.7\textwidth]{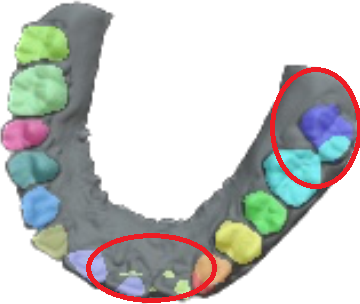}  
    \end{minipage}
    \begin{minipage}{0.15\textwidth} 
    \includegraphics[width=0.9\textwidth,height=0.7\textwidth]{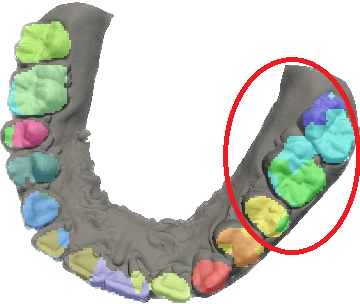}  
    \end{minipage}
    \begin{minipage}{0.15\textwidth} 
    \includegraphics[width=0.9\textwidth,height=0.7\textwidth]{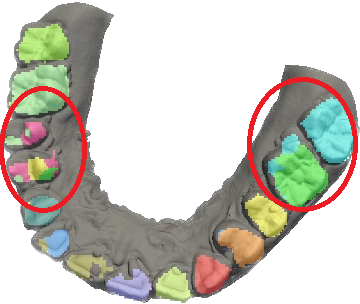}  
    \end{minipage}
    \begin{minipage}{0.15\textwidth} 
    \includegraphics[width=0.9\textwidth,height=0.7\textwidth]{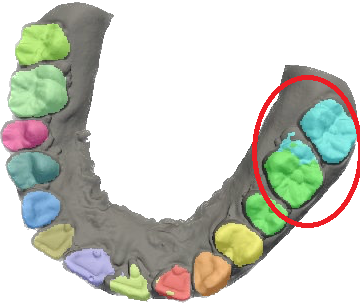}  
    \end{minipage}
    \begin{minipage}{0.15\textwidth} 
    \includegraphics[width=0.9\textwidth,height=0.7\textwidth]{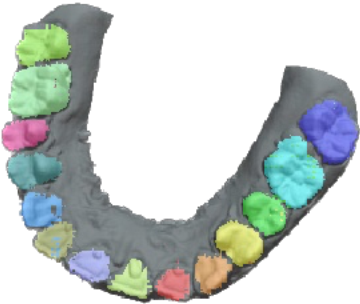}  
    \end{minipage}
    \begin{minipage}{0.15\textwidth} 
    \includegraphics[width=0.9\textwidth,height=0.7\textwidth]{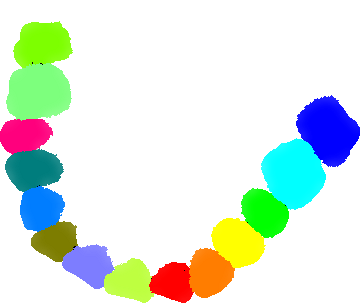}  
    \end{minipage}
    
    \end{minipage}
    \caption{The visual comparison of segmentation results (o.o.d test), as well as the corresponding ground truth.}
    \label{table:case}
\end{table*}

%% file: table/ablation.tex
\begin{table*}[ht]
\centering
\tiny
\resizebox{0.7\linewidth}{!}{
\begin{tabular}{c| c c c c c c| c c}
\toprule
\multicolumn{1}{c}{\multirow{2}{*}{\bf {Method}}} & \multicolumn{6}{c}{\bf {Module}}   & \multicolumn{1}{c}{\multirow{2}{*}{\bf {i.i.d. test}}}	 &  \multicolumn{1}{c}{\multirow{2}{*}{\bf {o.o.d. test}}}     \\
\cmidrule(lr){2-7}
\multicolumn{1}{c}{} & \multicolumn{1}{c}{Permute-UP} & \multicolumn{1}{c}{Bilinear-UP} & \multicolumn{1}{c}{Cross-Gate} & \multicolumn{1}{c}{Cross-ATT} & \multicolumn{1}{c}{MSA} & \multicolumn{1}{c}{APK} &\multicolumn{1}{c}{} &\multicolumn{1}{c}{}\\
\midrule
(a)  & \ding{55}  & \checkmark & \checkmark  &\ding{55}  & \checkmark & \checkmark   & 0.72 & 0.67 \\
(b)  & \checkmark & \ding{55}  & \ding{55}   &\checkmark & \checkmark & \checkmark   & 0.89 & 0.80 \\
(c)  & \checkmark & \ding{55}  & \checkmark  &\ding{55}  & \ding{55}  & \checkmark   & 0.87 & 0.73 \\
(d)  & \checkmark & \ding{55}  & \checkmark  &\ding{55}  & \checkmark & \ding{55}    & 0.89 & 0.76 \\
(e)  & \checkmark & \checkmark & -           & -         & \ding{55}  & \ding{55}    & 0.79 & 0.70 \\
(f)  & \checkmark & \ding{55}  & \checkmark  &\ding{55}  & \checkmark & \checkmark &\textbf{0.91} & \textbf{0.84} \\
\bottomrule
\end{tabular}%
}
\caption{Analysis of the effectiveness of each module.}
\label{table:effectiveness}
\end{table*}



%% file: sec/5_conclusion.tex
\section{Conclusion}
In this paper, we study the 2D image segmentation. To address the gap in research in this field, we create an open-source dataset called IO150k, which covers a wide range of dental malformations and is intended to serve as a resource for future research. 
Furthermore, starting from the particularity of the dental segmentation, we design TeethSEG, which surpasses the performance of the state-of-the-art segmentation models. This model includes two modules: Multi-Scale Aggregation (MSA) block and Anthropic Prior Knowledge (APK) layer. The former effectively aggregates the visual semantics into class
embeddings at different scales, and the latter imitates the principle of orthodontists to identify teeth. To realize MSA and APK, we developed a cross/self-gating mechanism for efficient deep feature interaction, as well as a permutation-based upscaler to generate clear segmentation edges and maintain local information in image patch embeddings. 
Experiments conducted in this paper demonstrate the effectiveness of our model and indicate that pretraining on plaster models can facilitate the segmentation of intra-oral images, which has the potential to assist large-scale epidemiological screenings and self-inspections.

%% file: sec/6_append.tex
\appendix
\renewcommand{\appendixname}{Appendix~\Alph{section}}
\cleardoublepage

\begin{figure*}[ht]
    \centering
    \includegraphics[width=1\textwidth]{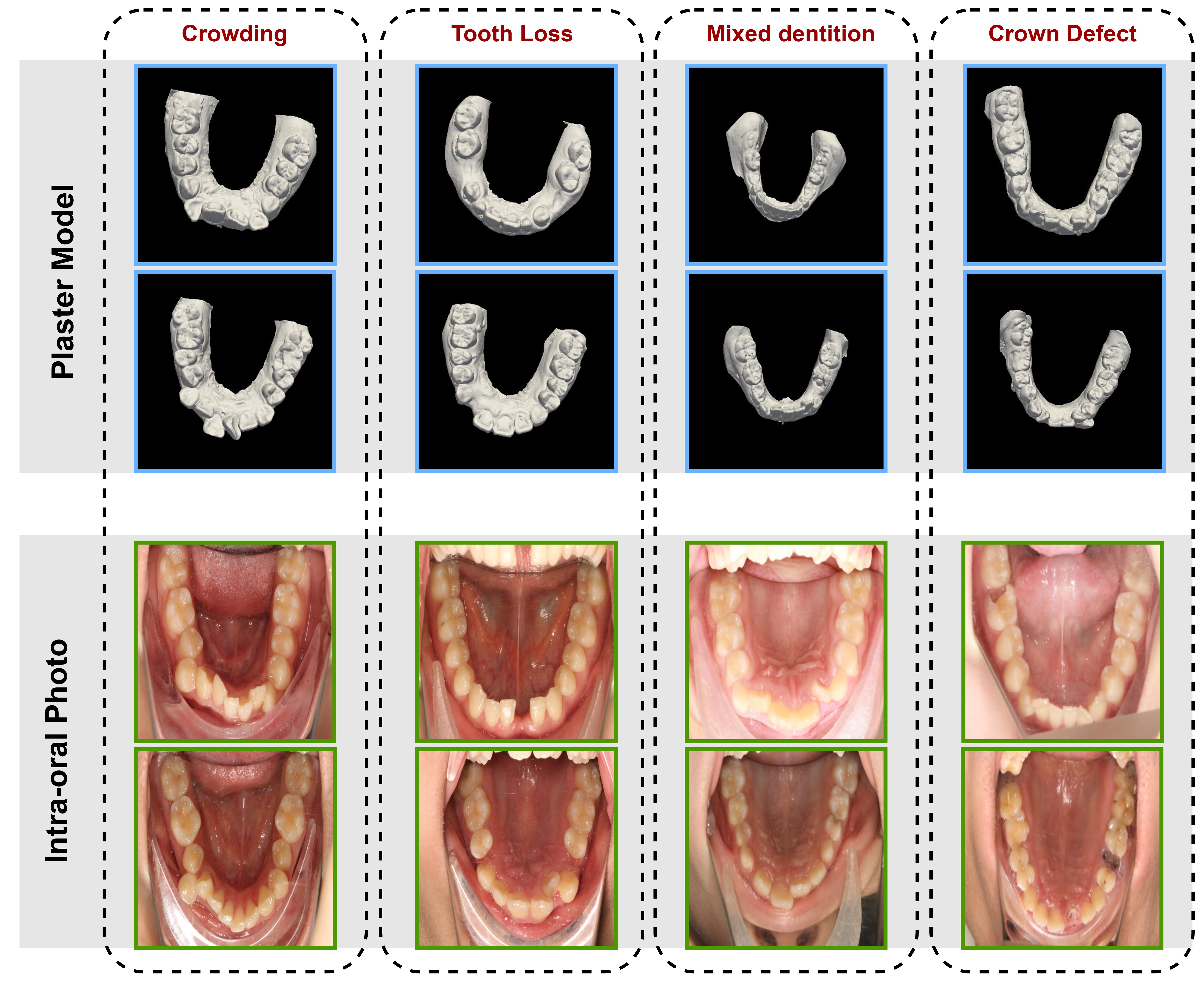}
    \caption{Visualization of Plaster70K and RGB0.8K}
    \label{fig:data-visual}
\end{figure*}
\begin{table*}[ht]
\centering
    \resizebox{0.7\textwidth}{!}{%
    \begin{tabular}{l|c c c | c c c c}
    \toprule
    \multirow{2}{*}{Name} &\multicolumn{3}{c|}{Split} &\multicolumn{4}{c}{dental malformations} \\
    &Train &Val &Test &Crowding &Tooth Loss &Mixed dentition &Crown Defect\\
    \midrule
    Challenge80K    &60k   &10k    &10k    &-           &-         &-          &-    \\

    Plaster70K      &50k   &10k    &10k    &44.1\%     &19.4\%     &4.5\%      &4.6\%\\

    RGB0.8K         &0.6k  &0.1k   &0.1k   &88.8\%     &8.4\%      &14.3\%     &2.5\%\\
    \bottomrule
    \end{tabular}
    }   
    \caption{Dataset statistics over the three subsets of IO150K.}
    \label{table:statistic}
\end{table*}

\section{Dataset Statistics}
We summarize the dataset statistics over the three subsets of IO150K in Table \ref{table:statistic} (we omit the statistic of dental malformations for Challenge80K, which is generated from previous open-sourced \cite{ben20233dteethseg}) and visualize some examples from our collected Plaster70K and RGB0.8K in Figure \ref{fig:data-visual}.

\clearpage

\input{table/ood}

\input{table/rgb}

\begin{figure}[ht]
    \centering
    \includegraphics[width=1\linewidth]{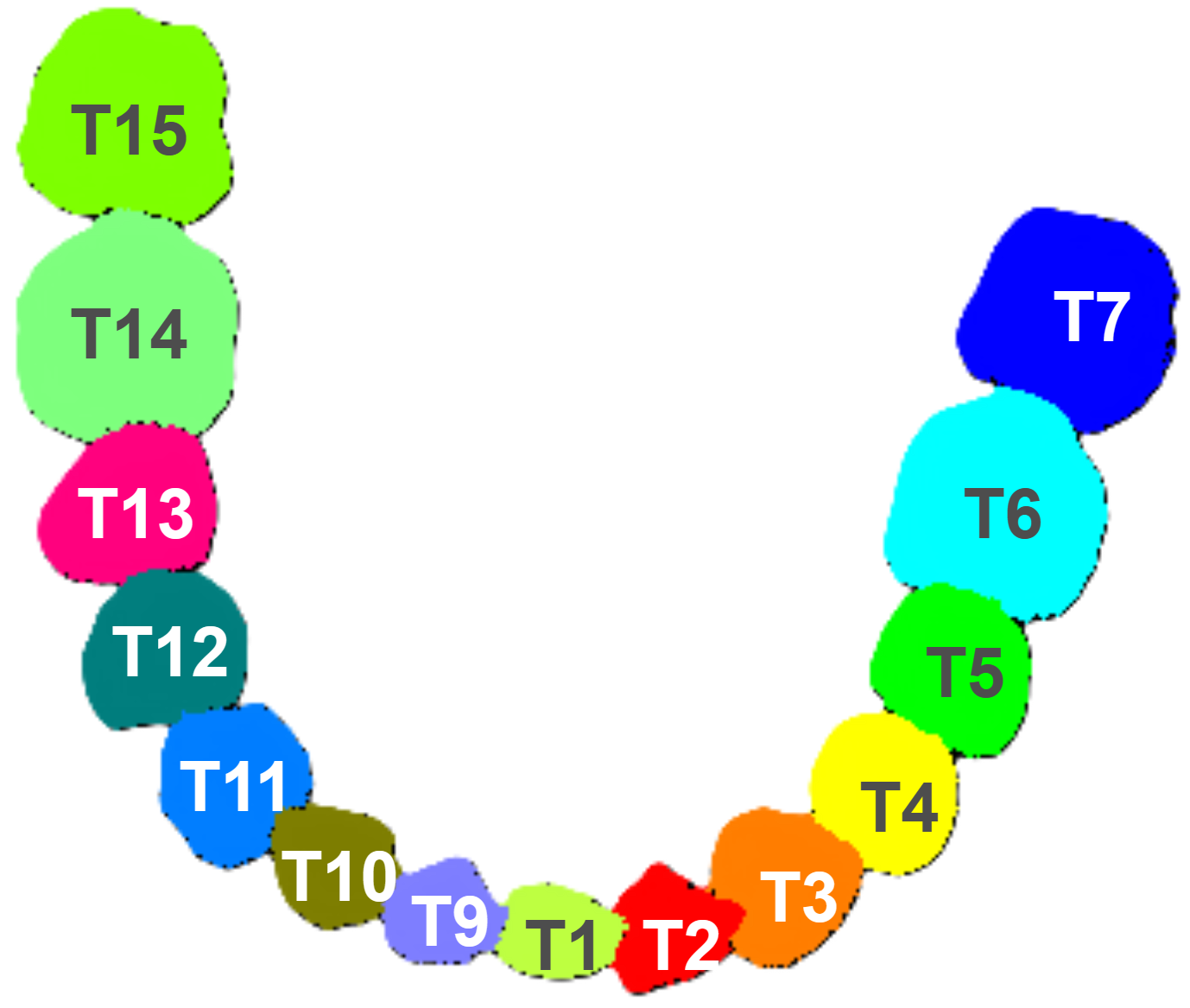}
    \caption{Classes Definition and Attention Mask in APK.}
    \label{fig:class}
\end{figure}
\begin{figure}[ht]
    \centering
    \includegraphics[width=0.9\linewidth]{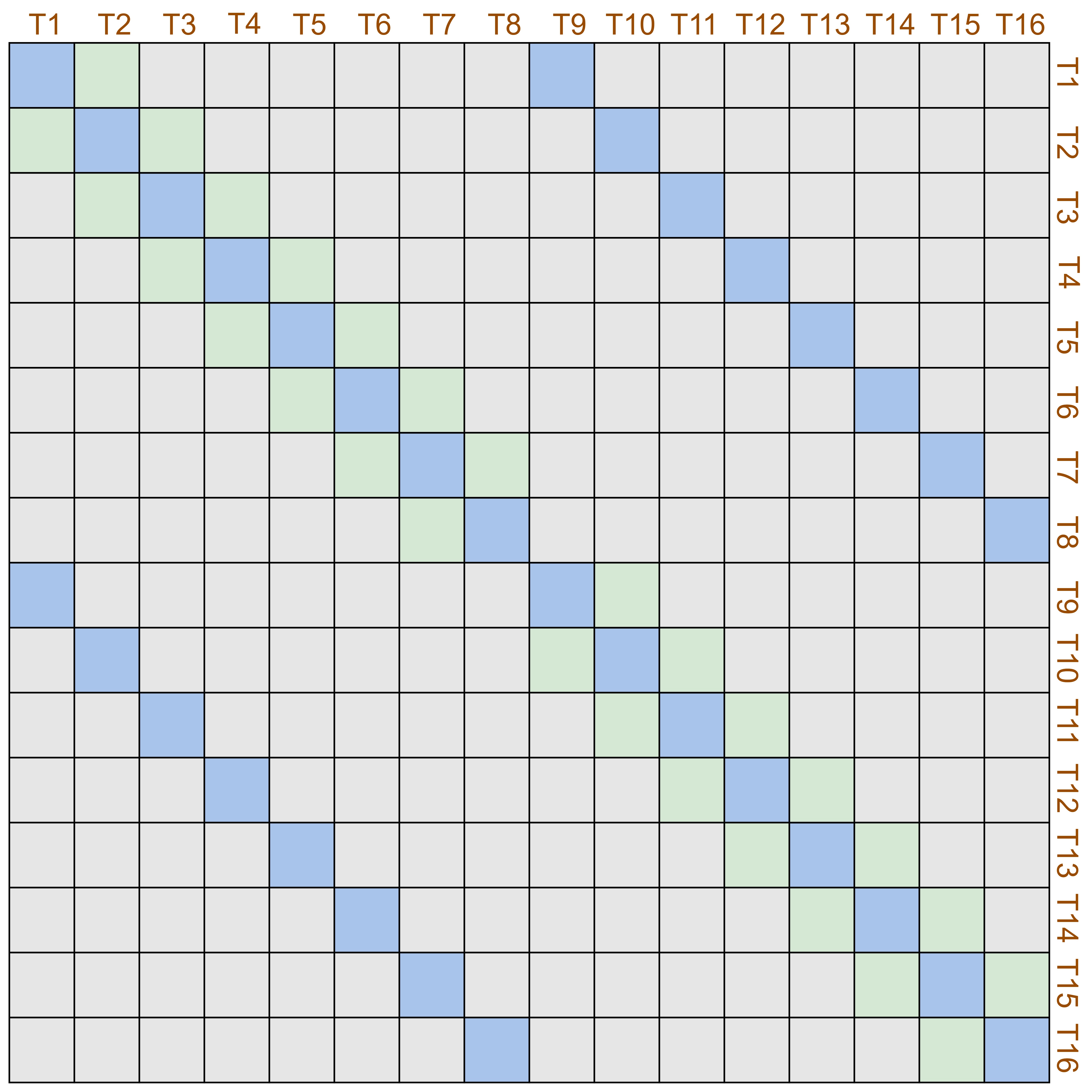}
    \caption{The attention mask of the APK layer.}
    \label{fig:mask}
\end{figure}

\section{The Classes Definition and The Attention Mask for The APK Layer}
The pre-defined order of Tooth IDs is presented in Figure \ref{fig:class}. In Figure \ref{fig:mask}, we visualize the attention mask that only enables interaction between adjacent (green) and contralateral (blue) teeth. We set the attention score of "$-inf$" for the locations with a gray color in the mask and $0$ for the locations with green or blue.

\section{Supplementary Results of The O.O.D. Test and The RGB Test.}
We provide the detailed quantitative results of each tooth (Table \ref{tab:ood_full}, Table \ref{tab:rgb}) and visualizations (Table \ref{table:case_full}, Table \ref{table:rgb_case}) for the o.o.d. test and RGB test. We find previous methods can still provide clear and appropriate segmentation boundaries but fail to provide accurate tooth IDs in complex situations of orthodontic treatment, especially in the RGB test.

\input{table/show_full}
\input{table/show_rgb}

\clearpage

\section{Ablations on Hyper-parameters}

\begin{table*}[ht]
\centering
\tiny
\resizebox{0.95\linewidth}{!}{
\begin{tabular}{c| c c c c c | c}
\toprule
\multirow{2}{*}{Variant}   &\multirow{2}{*}{Res.}   &Attention Layers   &Embed Dim.     &MSA Blocks         &Naive Upscalers     &\multirow{2}{*}{mIoU}\\
                           &                        &$M$                &$D$            &$N_{\rm{MSA}}$     &$N_{\rm{Up}}$      &  \\
\midrule
(a)                        &$224\times224$          &3                  &768            &3                  &2                  &0.83\\
\midrule
(b)                        &$336\times336$          &1                  &768            &3                  &2                  &0.82\\
(c)                        &$336\times336$          &2                  &768            &3                  &2                  &0.85\\
\midrule
(d)                        &$336\times336$          &3                  &256            &3                  &2                  &0.79\\
(e)                        &$336\times336$          &3                  &512            &3                  &2                  &0.89\\
\midrule
(f)                        &$336\times336$          &3                  &768            &1                  &4                  &0.81\\
(g)                        &$336\times336$          &3                  &768            &2                  &3                  &0.89\\
\midrule
(h)                        &$336\times336$          &3                  &768            &3                  &2                  &\textbf{0.91}\\
\bottomrule
\end{tabular}%
}
\caption{Ablations on Hyper-parameters.}
\label{table:more_ablations}
\end{table*}

We design a series of variants of TeethSEG to study the best choice of hyper-parameters. In Table \ref{table:more_ablations}, (h) presents the performance of the best variant in this paper, (a) studies the influence of resolution of inputs, (b) and (c) explore the influence of the number of the transformer layers for the shallow fusion, (d) and (e) investigate the impact intermediate dimension of embeddings, (f) and (g) study the consequence of the different number of stacked MSA Blocks. 

%% file: table/ood.tex
\begin{table*}[ht]
    \centering
    \label{tab:ood}
    \resizebox{\textwidth}{!}{%
    \begin{tabular}{l|c|ccccccccc cccccccc|c}
    \toprule
    Method          &Epoch  &T1     &T2     &T3     &T4     &T5     &T6     &T7     &T8     &T9     &T10    &T11    &T12    &T13    &T14    &T15    &T16    &mIoU\\
    \midrule
    DeepLab-v3      &30     &0.76   &0.72   &0.78   &0.85   &0.86   &0.81   &0.74   &NaN    &0.76   &0.75   &0.82   &0.84   &0.84   &0.87   &0.85   &NaN    &0.80\\
    Segformer       &30     &0.76   &0.73   &0.77   &0.85   &0.87   &0.80   &0.72   &NaN    &0.76   &0.76   &0.81   &0.87   &0.89   &0.90   &0.88   &NaN    &0.81\\
    Segmenter       &30     &0.63   &0.61   &0.63   &0.74   &0.70   &0.68   &0.59   &NaN    &0.62   &0.59   &0.68   &0.74   &0.70   &0.81   &0.78   &NaN    &0.68\\
    Swin-L          &30     &0.60   &0.58   &0.63   &0.69   &0.64   &0.67   &0.67   &NaN    &0.55   &0.52   &0.57   &0.63   &0.62   &0.73   &0.76   &NaN    &0.64\\
    SwinV2-G        &30     &0.52   &0.48   &0.52   &0.59   &0.57   &0.68   &0.67   &NaN    &0.49   &0.45   &0.53   &0.60   &0.66   &0.76   &0.76   &NaN    &0.60\\
    BeiT-B          &30     &0.78   &0.76   &0.77   &0.83   &0.74   &0.76   &0.77   &NaN    &0.79   &0.80   &0.82   &0.84   &0.76   &0.81   &0.82   &NaN    &0.78\\
    ViT-Adapter-L   &30     &0.78   &0.76   &0.77   &0.83   &0.74   &0.76   &0.77   &NaN    &0.79   &0.80   &0.82   &0.84   &0.76   &0.81   &0.82   &NaN    &0.79\\
    \midrule
    TeethSEG        &5      &0.80   &0.79   &0.82	&0.86   &0.86   &0.79   &0.71   &NaN    &0.78	&0.78	&0.83	&0.88	&0.90	&0.93	&0.91   &NaN    &0.84\\
    \bottomrule
    \end{tabular}
    }
    \caption{Tooth segmentation results (mIoU) compared with SOTA methods on the IO150K out-of-the-distribution (o.o.d.) test splits}
    \label{tab:ood_full}
\end{table*}

%% file: table/rgb.tex
\begin{table*}[ht]
    \centering

    \resizebox{\textwidth}{!}{%
    \begin{tabular}{l|c|ccccccccc cccccccc|c}
    \toprule
    Method          &Epoch  &T1     &T2     &T3     &T4     &T5     &T6     &T7     &T8     &T9     &T10    &T11    &T12    &T13    &T14    &T15    &T16    &mIoU\\
    \midrule
    DeepLab-v3       &30    &0.76   &0.72   &0.78   &0.85   &0.86   &0.81   &0.74   &NaN    &0.76   &0.75   &0.82   &0.84   &0.84   &0.87   &0.85   &NaN   &0.80\\
    Segformer       & 30    & 0.49  & 0.52  & 0.64  & 0.89  & 0.86  & 0.85  & 0.74  & NaN  & 0.45  & 0.50  & 0.65  & 0.84  & 0.83  & 0.84  & 0.57 & NaN  & 0.69 \\
    Segmenter       & 30    & 0.48  & 0.44  & 0.47  & 0.62  & 0.66  & 0.58  & 0.58  & NaN  & 0.52  & 0.49  & 0.49  & 0.60  & 0.63  & 0.57  & 0.55 & NaN  & 0.55 \\
    Swin-L          & 30    & 0.43  & 0.50  & 0.58  & 0.55  & 0.49  & 0.46  & 0.58  & NaN  & 0.41  & 0.51  & 0.43  & 0.35  & 0.35  & 0.51  & 0.59 & NaN  & 0.48 \\
    SwinV2-G        & 30    & 0.58  & 0.43  & 0.51  & 0.56  & 0.53  & 0.43  & 0.43  & NaN  & 0.60  & 0.37  & 0.44  & 0.42  & 0.41  & 0.42  & 0.32 & NaN  & 0.46 \\
    BeiT-B          & 30    & 0.61  & 0.48  & 0.46  & 0.49  & 0.53  & 0.39  & 0.35  & NaN  & 0.61  & 0.50  & 0.51  & 0.53  & 0.48  & 0.37  & 0.33 & NaN  & 0.47 \\
    ViT-Adapter-L   & 30    & 0.83  & 0.89  & 0.89  & 0.92  & 0.92  & 0.91  & 0.83  & NaN  & 0.85  & 0.88  & 0.86  & 0.92  & 0.92  & 0.89  & 0.72 & NaN  & 0.85 \\
    \midrule
    TeethSEG        &5      & 0.91  & 0.93  & 0.88	& 0.93   &0.94   &0.92   &0.87   &NaN    &0.86	&0.94	&0.90	&0.94	&0.93	&0.93	&0.91 & NaN & 0.91\\
    \bottomrule
    \end{tabular}
    }
    \caption{Tooth segmentation results (mIoU) compared with SOTA methods on the IO150K RGB test splits}
    \label{tab:rgb}
\end{table*}

%% file: table/show_full.tex
\begin{sidewaystable*}
    \begin{minipage}{1\textwidth} 
    
    \begin{minipage}{0.105\textwidth} 
    \centering
    DeepLab-v3
    \end{minipage}
    \begin{minipage}{0.105\textwidth} 
    \centering
    Segformer
    \end{minipage}
    \begin{minipage}{0.105\textwidth} 
    \centering
    Segmenter
    \end{minipage}
    \begin{minipage}{0.105\textwidth} 
    \centering
    BeiT-B
    \end{minipage}
    \begin{minipage}{0.105\textwidth} 
    \centering
    Swin-L
    \end{minipage}
    \begin{minipage}{0.105\textwidth} 
    \centering
    SwinV2-G
    \end{minipage}
    \begin{minipage}{0.105\textwidth} 
    \centering
    ViT-Adapter-L
    \end{minipage}
    \begin{minipage}{0.105\textwidth} 
    \centering
    TeethSEG
    (Ours)
    \end{minipage}
    \begin{minipage}{0.105\textwidth} 
    \centering
    Ground Truth
    \end{minipage}
    
    \end{minipage}
    \par
    \vspace{0.5cm}
    \begin{minipage}{1\textwidth} 
    
    \begin{minipage}{0.105\textwidth} 
    \includegraphics[width=1\textwidth,height=0.9\textwidth]{figure/imgs/LA108_38/fcn_cropped.png}  
    \end{minipage}
    \begin{minipage}{0.105\textwidth} 
    \includegraphics[width=1\textwidth,height=0.9\textwidth]{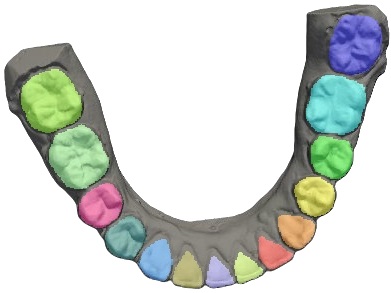}  
    \end{minipage}
    \begin{minipage}{0.105\textwidth} 
    \includegraphics[width=1\textwidth,height=0.9\textwidth]{figure/imgs/LA108_38/segformer.jpg}  
    \end{minipage}
    \begin{minipage}{0.105\textwidth} 
    \includegraphics[width=1\textwidth,height=0.9\textwidth]{figure/imgs/LA108_38/beit_cropped.png}  
    \end{minipage}
    \begin{minipage}{0.105\textwidth} 
    \includegraphics[width=1\textwidth,height=0.9\textwidth]{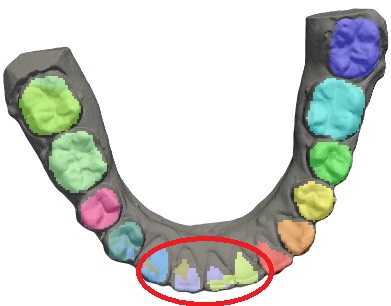}  
    \end{minipage}
    \begin{minipage}{0.105\textwidth} 
    \includegraphics[width=1\textwidth,height=0.9\textwidth]{figure/imgs/LA108_38/swinV2_cropped.png}  
    \end{minipage}
    \begin{minipage}{0.105\textwidth} 
    \includegraphics[width=1\textwidth,height=0.9\textwidth]{figure/imgs/LA108_38/adapter_cropped.png}  
    \end{minipage}
    \begin{minipage}{0.105\textwidth} 
    \includegraphics[width=1\textwidth,height=0.9\textwidth]{figure/imgs/LA108_38/teethSEG_cropped.png}  
    \end{minipage}
    \begin{minipage}{0.105\textwidth} 
    \includegraphics[width=1\textwidth,height=0.9\textwidth]{figure/imgs/LA108_38/gt_cropped.png}  
    \end{minipage}
    
    \end{minipage}
    \par
    \vspace{1.5cm}
    \begin{minipage}{1\textwidth} 
    
    \begin{minipage}{0.105\textwidth} 
    \includegraphics[width=1\textwidth,height=0.9\textwidth]{figure/imgs/LB201_16/fcn_cropped.png}  
    \end{minipage}
    \begin{minipage}{0.105\textwidth} 
    \includegraphics[width=1\textwidth,height=0.9\textwidth]{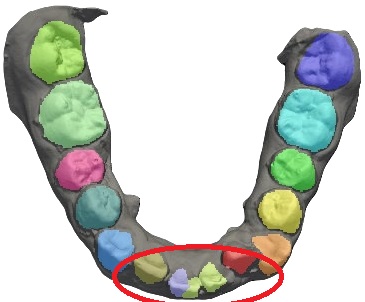}  
    \end{minipage}
    \begin{minipage}{0.105\textwidth} 
    \includegraphics[width=1\textwidth,height=0.9\textwidth]{figure/imgs/LB201_16/segformer.jpg}  
    \end{minipage}
    \begin{minipage}{0.105\textwidth} 
    \includegraphics[width=1\textwidth,height=0.9\textwidth]{figure/imgs/LB201_16/beit_cropped.png}  
    \end{minipage}
    \begin{minipage}{0.105\textwidth} 
    \includegraphics[width=1\textwidth,height=0.9\textwidth]{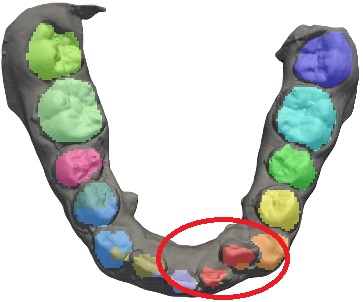}  
    \end{minipage}
    \begin{minipage}{0.105\textwidth} 
    \includegraphics[width=1\textwidth,height=0.9\textwidth]{figure/imgs/LB201_16/swinV2_cropped.png}  
    \end{minipage}
    \begin{minipage}{0.105\textwidth} 
    \includegraphics[width=1\textwidth,height=0.9\textwidth]{figure/imgs/LB201_16/adapter_cropped.png}  
    \end{minipage}
    \begin{minipage}{0.105\textwidth} 
    \includegraphics[width=1\textwidth,height=0.9\textwidth]{figure/imgs/LB201_16/teethSEG_cropped.png}  
    \end{minipage}
    \begin{minipage}{0.105\textwidth} 
    \includegraphics[width=1\textwidth,height=0.9\textwidth]{figure/imgs/LB201_16/gt_cropped.png}  
    \end{minipage}

    \end{minipage}
    \par
    \vspace{1.5cm}
    \begin{minipage}{1\textwidth} 
    
    \begin{minipage}{0.105\textwidth} 
    \includegraphics[width=1\textwidth,height=0.9\textwidth]{figure/imgs/UA199_51/fcn_cropped.png}  
    \end{minipage}
    \begin{minipage}{0.105\textwidth} 
    \includegraphics[width=1\textwidth,height=0.9\textwidth]{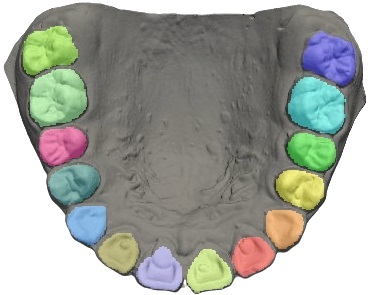}  
    \end{minipage}
    \begin{minipage}{0.105\textwidth} 
    \includegraphics[width=1\textwidth,height=0.9\textwidth]{figure/imgs/UA199_51/segformer.jpg}  
    \end{minipage}
    \begin{minipage}{0.105\textwidth} 
    \includegraphics[width=1\textwidth,height=0.9\textwidth]{figure/imgs/UA199_51/beit_cropped.png}  
    \end{minipage}
    \begin{minipage}{0.105\textwidth} 
    \includegraphics[width=1\textwidth,height=0.9\textwidth]{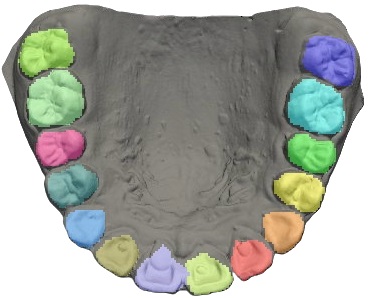}  
    \end{minipage}
    \begin{minipage}{0.105\textwidth} 
    \includegraphics[width=1\textwidth,height=0.9\textwidth]{figure/imgs/UA199_51/swinV2_cropped.png}  
    \end{minipage}
    \begin{minipage}{0.105\textwidth} 
    \includegraphics[width=1\textwidth,height=0.9\textwidth]{figure/imgs/UA199_51/adapter_cropped.png}  
    \end{minipage}
    \begin{minipage}{0.105\textwidth} 
    \includegraphics[width=1\textwidth,height=0.9\textwidth]{figure/imgs/UA199_51/teethSEG_cropped.png}  
    \end{minipage}
    \begin{minipage}{0.105\textwidth} 
    \includegraphics[width=1\textwidth,height=0.9\textwidth]{figure/imgs/UA199_51/gt_cropped.png}  
    \end{minipage}   
    
    \end{minipage}
    \par
    \vspace{1.5cm}
    \begin{minipage}{1\textwidth} 
    
    \begin{minipage}{0.105\textwidth} 
    \includegraphics[width=1\textwidth,height=0.9\textwidth]{figure/imgs/UB108_70/fcn_cropped.png}  
    \end{minipage}
    \begin{minipage}{0.105\textwidth} 
    \includegraphics[width=1\textwidth,height=0.9\textwidth]{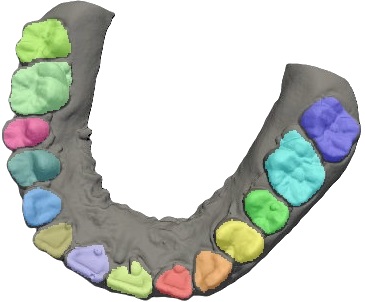}  
    \end{minipage}
    \begin{minipage}{0.105\textwidth} 
    \includegraphics[width=1\textwidth,height=0.9\textwidth]{figure/imgs/UB108_70/segformer.jpg}  
    \end{minipage}
    \begin{minipage}{0.105\textwidth} 
    \includegraphics[width=1\textwidth,height=0.9\textwidth]{figure/imgs/UB108_70/beit_cropped.png}  
    \end{minipage}
    \begin{minipage}{0.105\textwidth} 
    \includegraphics[width=1\textwidth,height=0.9\textwidth]{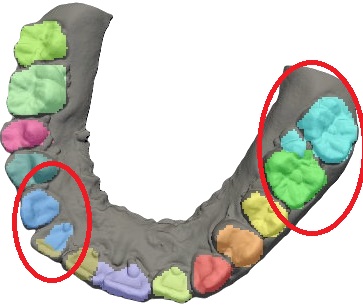}  
    \end{minipage}
    \begin{minipage}{0.105\textwidth} 
    \includegraphics[width=1\textwidth,height=0.9\textwidth]{figure/imgs/UB108_70/swinV2_cropped.png}  
    \end{minipage}
    \begin{minipage}{0.105\textwidth} 
    \includegraphics[width=1\textwidth,height=0.9\textwidth]{figure/imgs/UB108_70/adapter_cropped.png}  
    \end{minipage}
    \begin{minipage}{0.105\textwidth} 
    \includegraphics[width=1\textwidth,height=0.9\textwidth]{figure/imgs/UB108_70/teethSEG_cropped.png}  
    \end{minipage}
    \begin{minipage}{0.105\textwidth} 
    \includegraphics[width=1\textwidth,height=0.9\textwidth]{figure/imgs/UB108_70/gt_cropped.png}  
    \end{minipage}
    
    \end{minipage}
    \vspace{0.5cm}
    \caption{The visual comparison of segmentation results (o.o.d test), as well as the corresponding ground truth.}
    \label{table:case_full}
\end{sidewaystable*}

%% file: table/show_rgb.tex
\begin{sidewaystable*}
    \begin{minipage}{1\textwidth} 
    
    \begin{minipage}{0.105\textwidth} 
    \centering
    DeepLab-v3
    \end{minipage}
    \begin{minipage}{0.105\textwidth} 
    \centering
    Segformer
    \end{minipage}
    \begin{minipage}{0.105\textwidth} 
    \centering
    Segmenter
    \end{minipage}
    \begin{minipage}{0.105\textwidth} 
    \centering
    BeiT-B
    \end{minipage}
    \begin{minipage}{0.105\textwidth} 
    \centering
    Swin-L
    \end{minipage}
    \begin{minipage}{0.105\textwidth} 
    \centering
    SwinV2-G
    \end{minipage}
    \begin{minipage}{0.105\textwidth} 
    \centering
    ViT-Adapter-L
    \end{minipage}
    \begin{minipage}{0.105\textwidth} 
    \centering
    TeethSEG
    (Ours)
    \end{minipage}
    \begin{minipage}{0.105\textwidth} 
    \centering
    Ground Truth
    \end{minipage}
    
    \end{minipage}
    \par
    \vspace{0.5cm}
    \begin{minipage}{1\textwidth} 
    
    \begin{minipage}{0.105\textwidth} 
    \includegraphics[width=1\textwidth,height=1\textwidth]{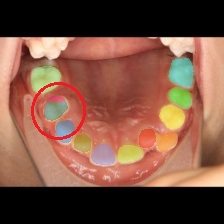}  
    \end{minipage}
    \begin{minipage}{0.105\textwidth} 
    \includegraphics[width=1\textwidth,height=1\textwidth]{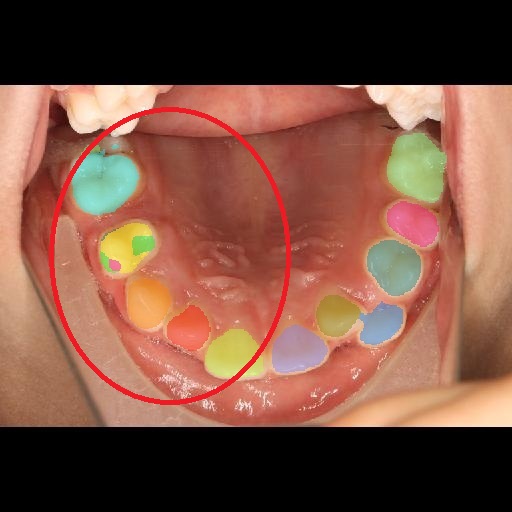}  
    \end{minipage}
    \begin{minipage}{0.105\textwidth} 
    \includegraphics[width=1\textwidth,height=1\textwidth]{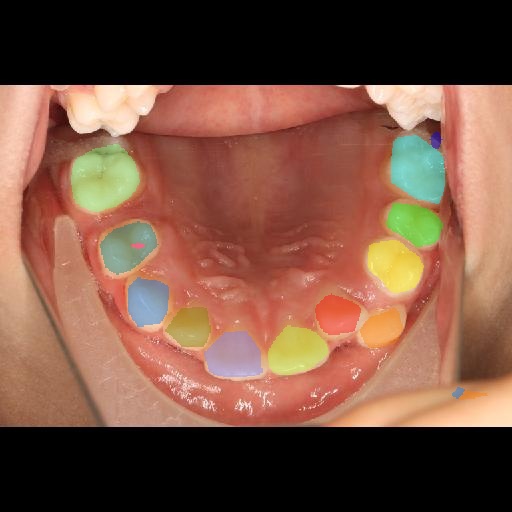}  
    \end{minipage}
    \begin{minipage}{0.105\textwidth} 
    \includegraphics[width=1\textwidth,height=1\textwidth]{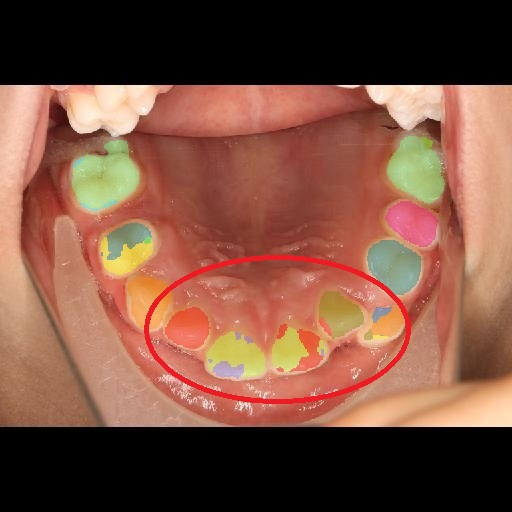}  
    \end{minipage}
    \begin{minipage}{0.105\textwidth} 
    \includegraphics[width=1\textwidth,height=1\textwidth]{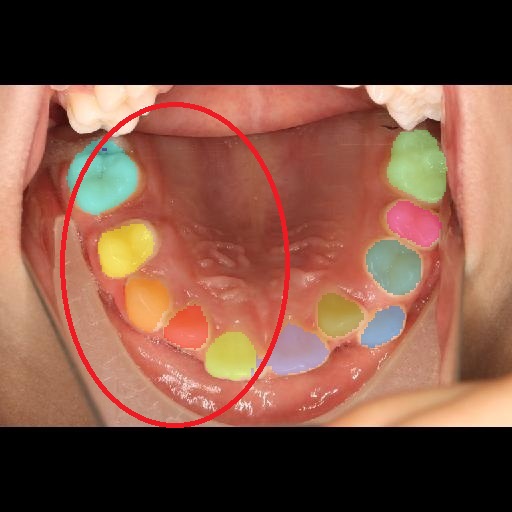}  
    \end{minipage}
    \begin{minipage}{0.105\textwidth} 
    \includegraphics[width=1\textwidth,height=1\textwidth]{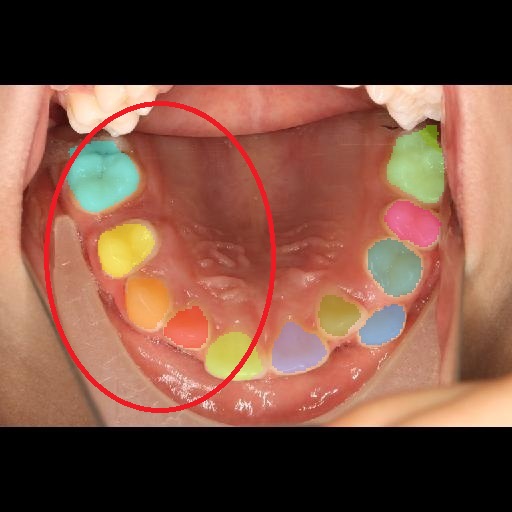}  
    \end{minipage}
    \begin{minipage}{0.105\textwidth} 
    \includegraphics[width=1\textwidth,height=1\textwidth]{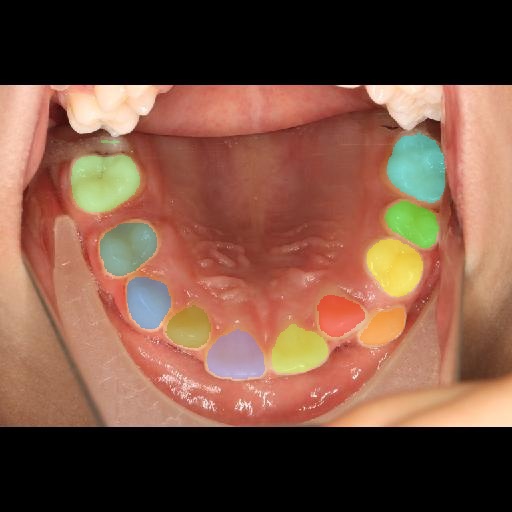}  
    \end{minipage}
    \begin{minipage}{0.105\textwidth} 
    \includegraphics[width=1\textwidth,height=1\textwidth]{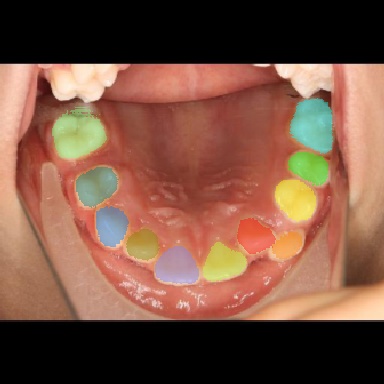}  
    \end{minipage}
    \begin{minipage}{0.105\textwidth} 
    \includegraphics[width=1\textwidth,height=1\textwidth]{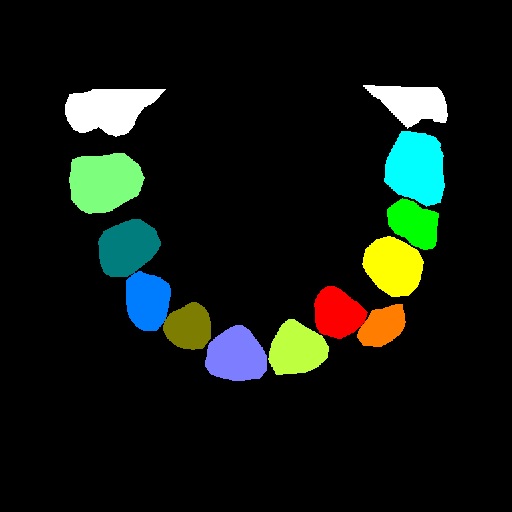}  
    \end{minipage}
    
    \end{minipage}
    \par
    \vspace{1.15cm}
    \begin{minipage}{1\textwidth} 
    
    \begin{minipage}{0.105\textwidth} 
    \includegraphics[width=1\textwidth,height=1\textwidth]{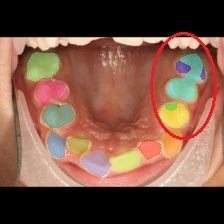}  
    \end{minipage}
    \begin{minipage}{0.105\textwidth} 
    \includegraphics[width=1\textwidth,height=1\textwidth]{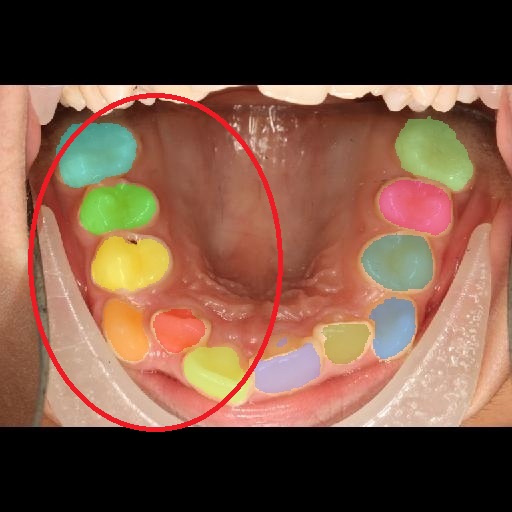}  
    \end{minipage}
    \begin{minipage}{0.105\textwidth} 
    \includegraphics[width=1\textwidth,height=1\textwidth]{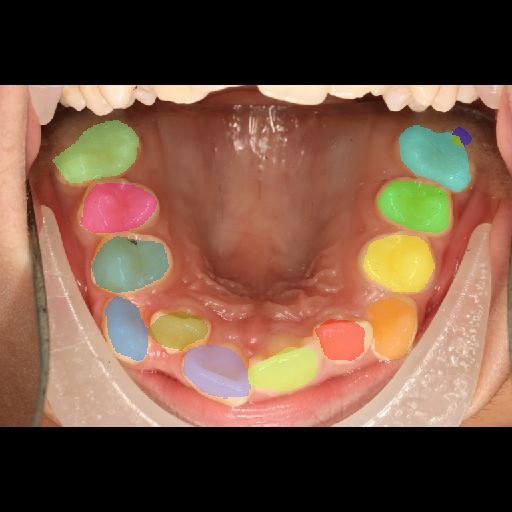}  
    \end{minipage}
    \begin{minipage}{0.105\textwidth} 
    \includegraphics[width=1\textwidth,height=1\textwidth]{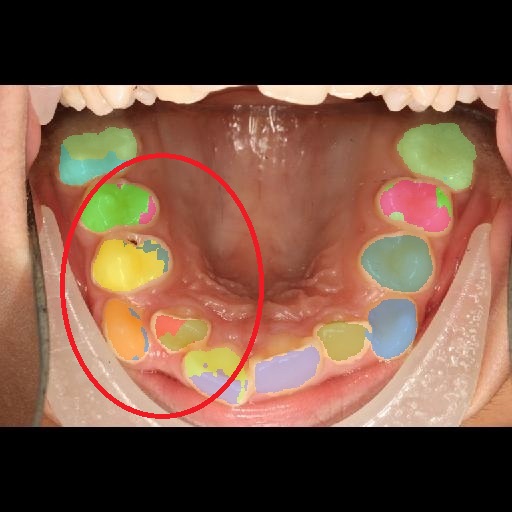}  
    \end{minipage}
    \begin{minipage}{0.105\textwidth} 
    \includegraphics[width=1\textwidth,height=1\textwidth]{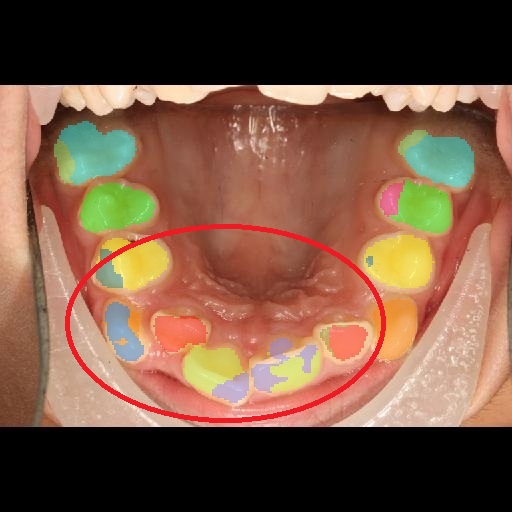}  
    \end{minipage}
    \begin{minipage}{0.105\textwidth} 
    \includegraphics[width=1\textwidth,height=1\textwidth]{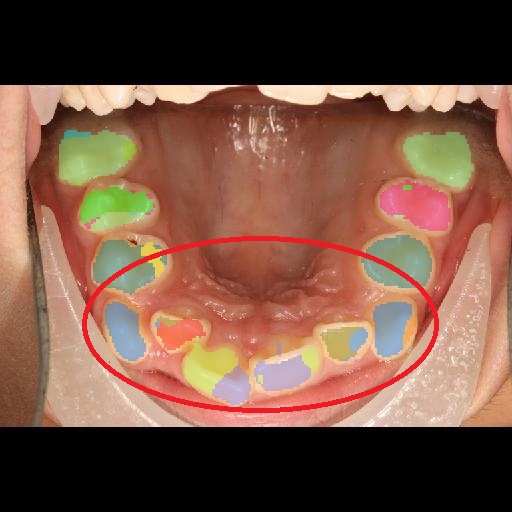}  
    \end{minipage}
    \begin{minipage}{0.105\textwidth} 
    \includegraphics[width=1\textwidth,height=1\textwidth]{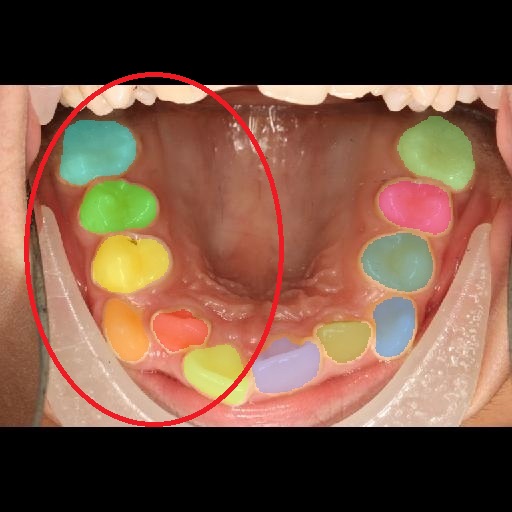}  
    \end{minipage}
    \begin{minipage}{0.105\textwidth} 
    \includegraphics[width=1\textwidth,height=1\textwidth]{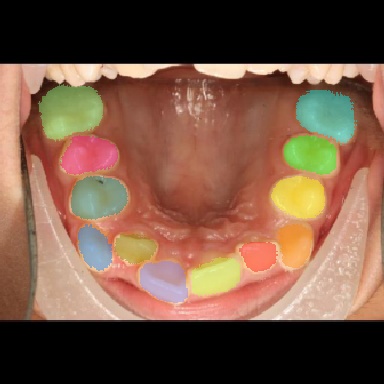}  
    \end{minipage}
    \begin{minipage}{0.105\textwidth} 
    \includegraphics[width=1\textwidth,height=1\textwidth]{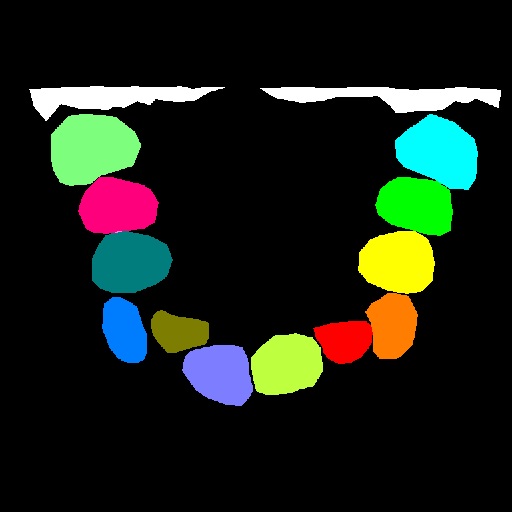}  
    \end{minipage}

    \end{minipage}
    \par
    \vspace{1.15cm}
    \begin{minipage}{1\textwidth} 
    
    \begin{minipage}{0.105\textwidth} 
    \includegraphics[width=1\textwidth,height=1\textwidth]{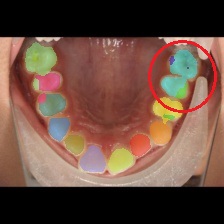}  
    \end{minipage}
    \begin{minipage}{0.105\textwidth} 
    \includegraphics[width=1\textwidth,height=1\textwidth]{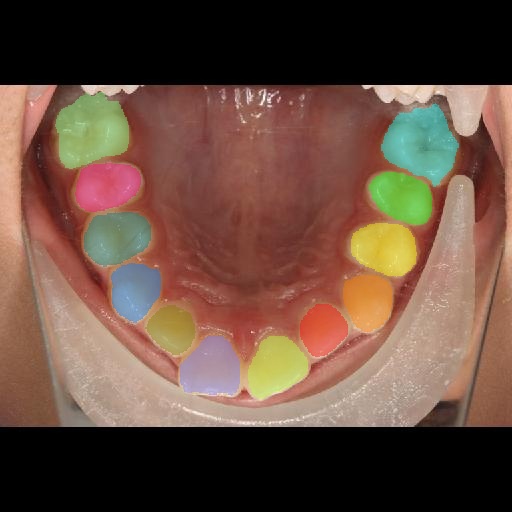}  
    \end{minipage}
    \begin{minipage}{0.105\textwidth} 
    \includegraphics[width=1\textwidth,height=1\textwidth]{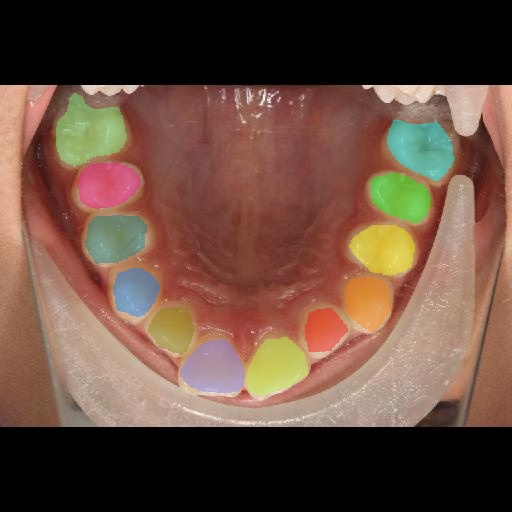}  
    \end{minipage}
    \begin{minipage}{0.105\textwidth} 
    \includegraphics[width=1\textwidth,height=1\textwidth]{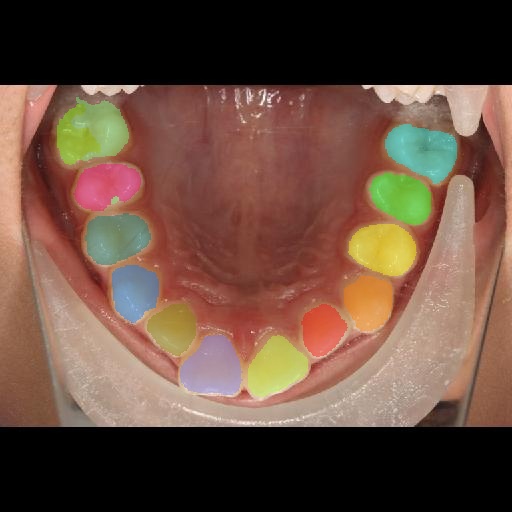}  
    \end{minipage}
    \begin{minipage}{0.105\textwidth} 
    \includegraphics[width=1\textwidth,height=1\textwidth]{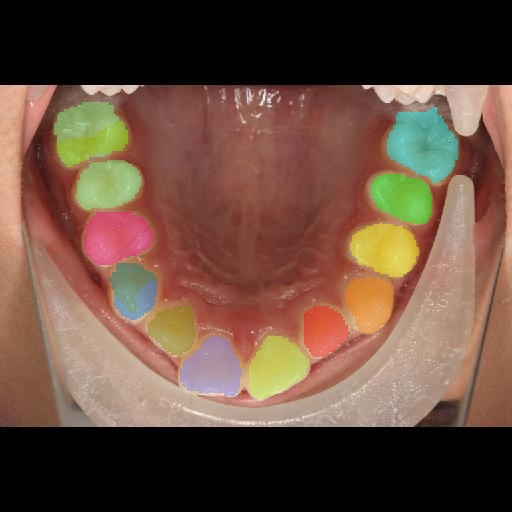}  
    \end{minipage}
    \begin{minipage}{0.105\textwidth} 
    \includegraphics[width=1\textwidth,height=1\textwidth]{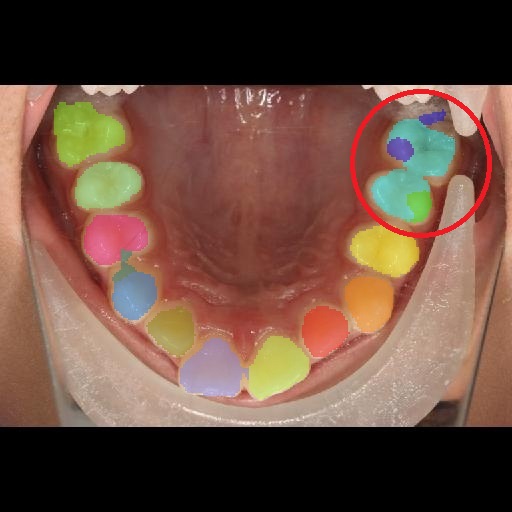}  
    \end{minipage}
    \begin{minipage}{0.105\textwidth} 
    \includegraphics[width=1\textwidth,height=1\textwidth]{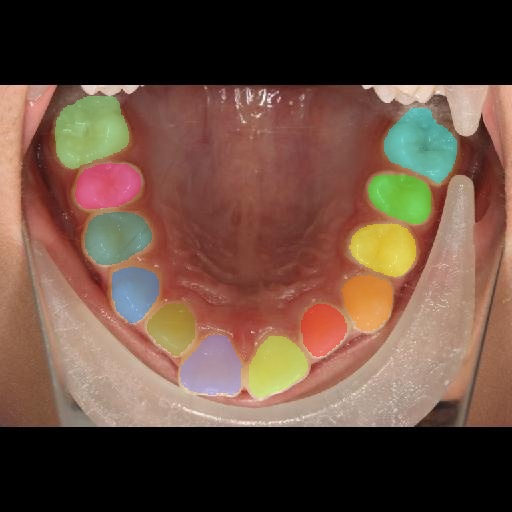}  
    \end{minipage}
    \begin{minipage}{0.105\textwidth} 
    \includegraphics[width=1\textwidth,height=1\textwidth]{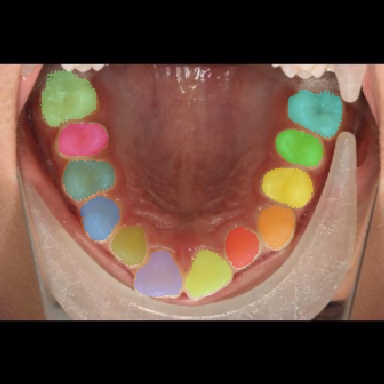}  
    \end{minipage}
    \begin{minipage}{0.105\textwidth} 
    \includegraphics[width=1\textwidth,height=1\textwidth]{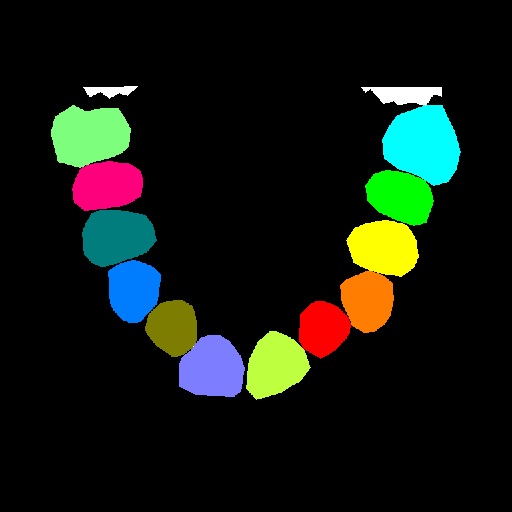}  
    \end{minipage}   
    
    \end{minipage}
    \par
    \vspace{1.15cm}
    \begin{minipage}{1\textwidth} 
    
    \begin{minipage}{0.105\textwidth} 
    \includegraphics[width=1\textwidth,height=1\textwidth]{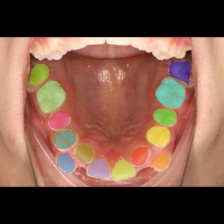}  
    \end{minipage}
    \begin{minipage}{0.105\textwidth} 
    \includegraphics[width=1\textwidth,height=1\textwidth]{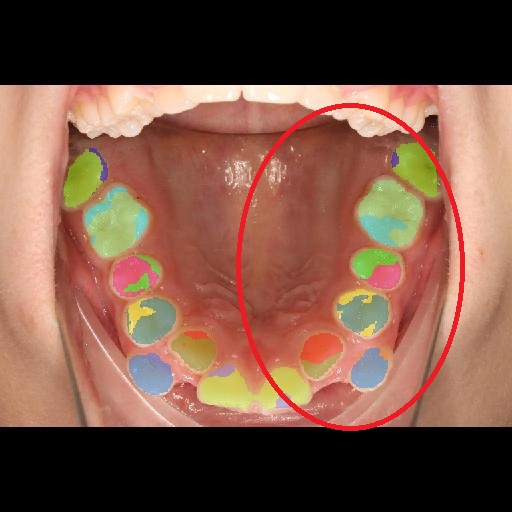}  
    \end{minipage}
    \begin{minipage}{0.105\textwidth} 
    \includegraphics[width=1\textwidth,height=1\textwidth]{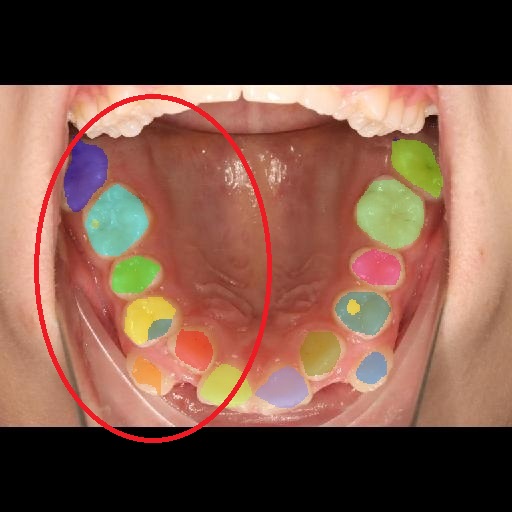}  
    \end{minipage}
    \begin{minipage}{0.105\textwidth} 
    \includegraphics[width=1\textwidth,height=1\textwidth]{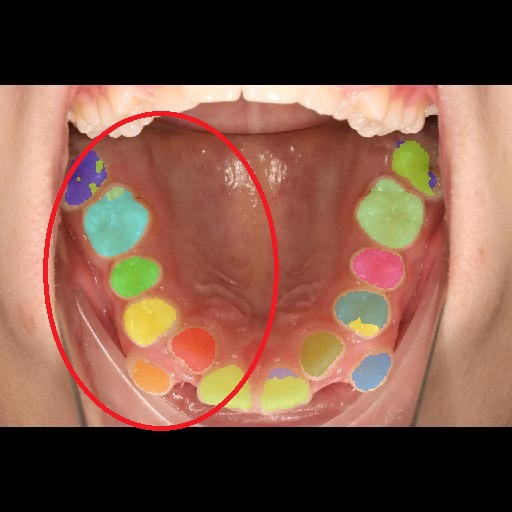}  
    \end{minipage}
    \begin{minipage}{0.105\textwidth} 
    \includegraphics[width=1\textwidth,height=1\textwidth]{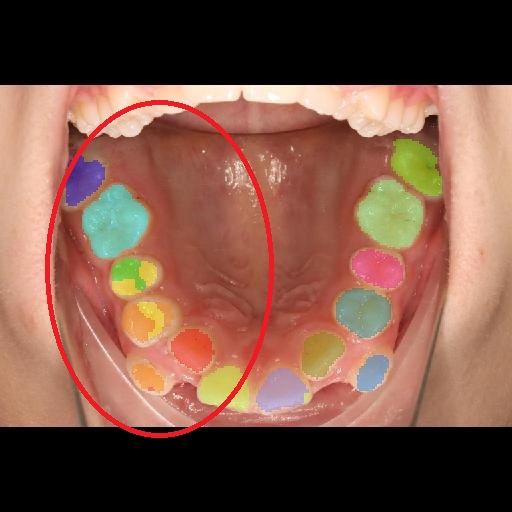}  
    \end{minipage}
    \begin{minipage}{0.105\textwidth} 
    \includegraphics[width=1\textwidth,height=1\textwidth]{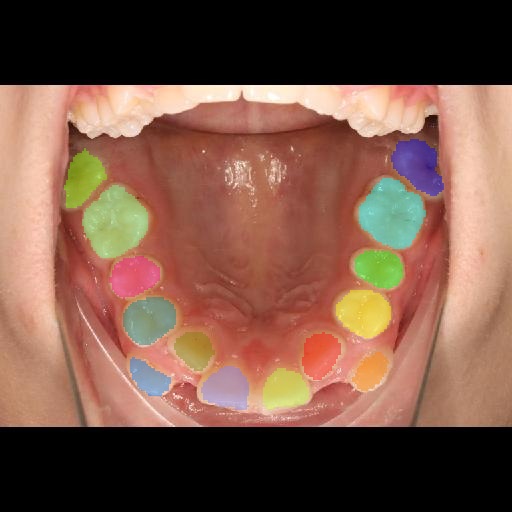}  
    \end{minipage}
    \begin{minipage}{0.105\textwidth} 
    \includegraphics[width=1\textwidth,height=1\textwidth]{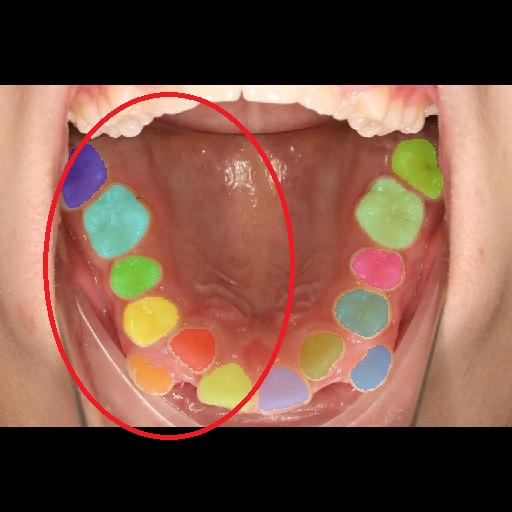}  
    \end{minipage}
    \begin{minipage}{0.105\textwidth} 
    \includegraphics[width=1\textwidth,height=1\textwidth]{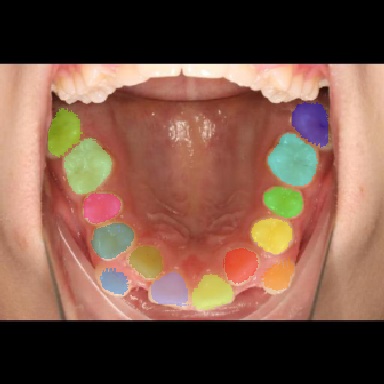}  
    \end{minipage}
    \begin{minipage}{0.105\textwidth} 
    \includegraphics[width=1\textwidth,height=1\textwidth]{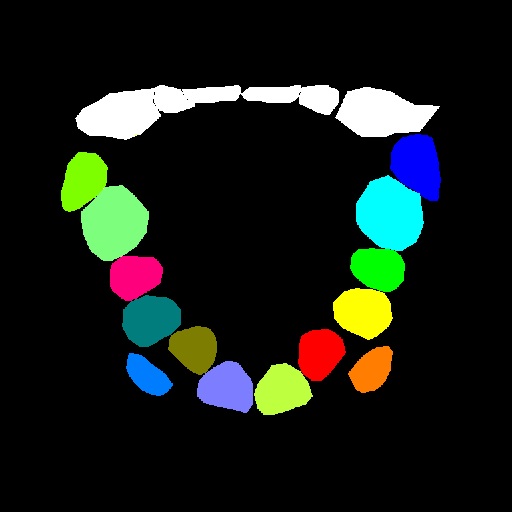}  
    \end{minipage}
    
    \end{minipage}
    \vspace{0.5cm}
    \caption{The visual comparison on RGB test. Previous methods fail to provide accurate tooth IDs in complex situations of orthodontic treatment.}
    \label{table:rgb_case}
\end{sidewaystable*}